\documentclass{article}

\usepackage{microtype}
\usepackage{graphicx}
\usepackage{subfigure}
\usepackage{booktabs} %
\usepackage{multirow}
\usepackage{makecell}
\usepackage{hyperref}
\usepackage{pifont}
\usepackage{svg}

\usepackage[accepted]{icml2024}

\usepackage{amsmath}
\usepackage{amssymb}
\usepackage{mathtools}
\usepackage{amsthm}
\usepackage{enumitem}
\usepackage[capitalize,noabbrev]{cleveref}

\usepackage{booktabs}
\usepackage{multirow}
\usepackage{graphicx}
\usepackage[normalem]{ulem}
\useunder{\uline}{\ul}{}

\theoremstyle{plain}

\theoremstyle{definition}

\theoremstyle{remark}

\usepackage[textsize=tiny]{todonotes}
\usepackage{hyperref} 

\icmltitlerunning{Behavior Generation with Latent Actions}

\begin{document}

\twocolumn[
\icmltitle{Behavior Generation with Latent Actions}

\icmlsetsymbol{equal}{*}

\begin{icmlauthorlist}

\icmlauthor{Seungjae Lee}{nyu,snu}
\icmlauthor{Yibin Wang}{nyu}
\icmlauthor{Haritheja Etukuru}{nyu}
\icmlauthor{H. Jin Kim}{snu,aiis}\\
\icmlauthor{Nur Muhammad Mahi Shafiullah}{equal,nyu}
\icmlauthor{Lerrel Pinto}{equal,nyu}
\end{icmlauthorlist}

\icmlaffiliation{snu}{Department of Aerospace Engineering, Seoul National University}
\icmlaffiliation{nyu}{New York University}
\icmlaffiliation{aiis}{Artificial Intelligence Institute of SNU  * Equal Advising}

\icmlcorrespondingauthor{Nur Muhammad Mahi Shafiullah}{\href{mailto:mahi@cs.nyu.edu}{mahi@cs.nyu.edu}}

\icmlkeywords{Machine Learning, ICML}

\vskip 0.3in
]

\newcommand{\xxnote}[3]{}
\renewcommand{\xxnote}[3]{\color{#2}{#1: #3}}
\newcommand{\LP}[1]{{\xxnote{LP}{red}{#1}}}
\newcommand{\MS}[1]{{\xxnote{MS}{blue}{#1}}}
\newcommand{\AR}[1]{{\xxnote{AR}{green}{#1}}}
\newcommand{\PL}[1]{{\xxnote{Peiqi}{orange}{#1}}}
\newcommand{\YO}[1]{{\xxnote{Yaswanth}{cyan}{#1}}}
\newcommand{\SJ}[1]{{\xxnote{Jay}{brown}{#1}}}
\newcommand{\method}{VQ-BeT}
\newcommand{\methodlong}{Vector-Quantized Behavior Transformer}

\printAffiliationsAndNotice{}  %

\begin{abstract}

Generative modeling of complex behaviors from labeled datasets has been a longstanding problem in decision-making. Unlike language or image generation, decision-making requires modeling actions -- continuous-valued vectors that are multimodal in their distribution, potentially drawn from uncurated sources, where  generation errors can compound in sequential prediction. A recent class of models called Behavior Transformers (BeT) addresses this by discretizing actions using k-means clustering to capture different modes. However, k-means struggles to scale for high-dimensional action spaces or long sequences, and lacks gradient information, and thus BeT suffers in modeling long-range actions. In this work, we present~\methodlong{} (\method{}), a versatile model for behavior generation that handles multimodal action prediction, conditional generation, and partial observations. \method{} augments BeT by tokenizing continuous actions with a hierarchical vector quantization module. Across seven environments including simulated manipulation, autonomous driving, and robotics, \method{} improves on state-of-the-art models such as BeT  and Diffusion Policies. Importantly, we demonstrate \method{}'s improved ability to capture behavior modes while accelerating inference speed $5\times$ over Diffusion Policies. Videos can be found \href{https://sjlee.cc/vq-bet/}{https://sjlee.cc/vq-bet/}

\end{abstract}

\section{Introduction}
\label{introduction}

The presently dominant paradigm in modeling human outputs, whether in language~\cite{achiam2023gpt}, image~\cite{podell2023sdxl}, audio~\cite{ziv2024masked}, or video~\cite{bar2024lumiere}, follows a similar recipe: collect a large in-domain dataset, use a large model that fits the dataset, and possibly as a cherry on top, improve the model output using some domain-specific feedback or datasets.
However, such a large, successful model for generating human or robot actions in embodied environments has been absent so far, and the issues are apparent. Action sequences are semantically diverse but temporally highly correlated, human behavior distributions are massively multi-modal and noisy, and the hard-and-fast grounding in the laws of physics means that unlike audio, language or video-generation, even the smallest discrepancies may cause a cascade of consequences that lead to catastrophic failures in as few as tens of timesteps \cite{ross2011reduction, rajaraman2020toward}.
The desiderata for a good model of behaviors and actions thus must contain the following abilities: to model long- and short-term dependencies, to capture and generate from diverse modes of behavior, and to replicate the learned behaviors precisely \cite{shafiullah2022behavior, chi2023diffusion}.

\begin{figure*}[t]
\begin{center}
\centerline{\includegraphics[width=.93\linewidth]{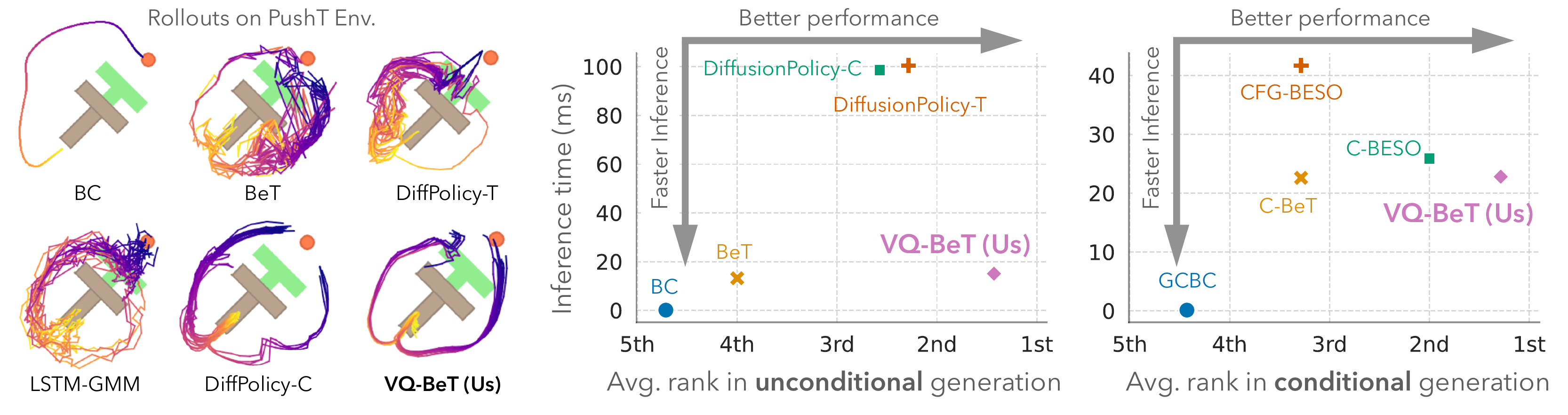}}
\vskip -0.1in
\caption{Qualitative and quantitative comparison between \method{} and relevant baselines. On the left, we can see trajectories generated by different algorithms while pushing a T-block to target, where ~\method{} generates smooth trajectories covering both modes. On the right, we show two plots comparing \method{} and relevant baselines on unconditional and goal-conditional behavior generation. The comparison axes are (x-axis) relative success represented by average rank on a suite of seven simulated tasks, and (y-axis) inference time.}
\label{fig:intro}
\vskip -0.3in
\end{center}
\end{figure*}

Prior work by \citep{shafiullah2022behavior} shows how transformers can capture the temporal dependencies well, and to some extent even capture the multi-modality in the data with clever tokenization.
However, that tokenziation relies on k-means clustering, a method typically based on an $\ell_2$ metric space that unfortunately does not scale to high-dimensional action spaces or temporally extended actions with lots of inter-dependencies.
More recent works have also used tools from generative modeling to address the problem of behavior modeling~\cite{pearce2023imitating, chi2023diffusion, zhao2023learning}, but issues remain, for example in high computational cost when scaling to long-horizons, or failing to express multi-modality during rollouts.

In this work, we propose~\methodlong{} (\method{}), which combines the long-horizon modeling capabilities of transformers with the expressiveness of vector-quantization %
to minimize the compute cost while maintaining high fidelity to the data.
We posit that a large part of the difficulty in behavior modeling comes from representing the continuous-valued, multi-modal action vectors.
A ready answer is learning discrete representations using vector quantization~\cite{van2017neural} used extensively to handle the output spaces in audio~\cite{dhariwal2020jukebox}, video~\cite{wu2021godiva}, and image~\cite{rombach2022high}. %
In particular, the performance of VQ-VAEs for generative tasks has been so strong that a lot of recent models that generate continuous values simply generate a latent vector in the VQ-space first before decoding or upsampling the result~\cite{ziv2024masked, bar2024lumiere, podell2023sdxl}.

\method{} is designed to be versatile, allowing it to be readily used in both conditional and unconditional generation, while being performative on problems ranging across simulated manipulation, autonomous driving, and real-robotics. Through extensive experiments across eight benchmark environments, we present the following experimental insights:

\begin{enumerate}
    \vspace{-2mm}
    \item \method{} achieves state-of-the-art (SOTA) performance on unconditional behavior generation outperforming BC, BeT, and diffusion policies in $5/7$ environments (Figure~\ref{fig:intro} middle). Quantitative metrics of entropy and qualitative visualizations indicate that this performance gain is due to better capture of multiple modes in behavior data (Figure~\ref{fig:intro} left).
    \item On conditional behavior generation, by simply specifying goals as input, \method{} achieves SOTA performance and improves upon GCBC, C-BeT, and BESO in $6/7$ environments (Figure~\ref{fig:intro} right).
    \item \method{} directly works on autonomous driving benchmarks such as nuScenes~\cite{caesar2020nuscenes}, matching and being comparable to task-specific SOTA methods. %
    \item \method{} is a single-pass model, and hence offers a $5\times$ speedup in simulation and $25\times$ on real-world robots over multi-pass models that use diffusion models. 
    \item \method{} scales to real-world robotic manipulation such as pick-and-placing objects and door closing, improving upon prior work by $73\%$ on long-horizon tasks.
\end{enumerate}

\section{Background and Preliminaries}
\label{background}
\subsection{Behavior cloning}
\label{sec:behavior_cloning}
Given a dataset of continuous-valued action and observation pairs $\mathcal{D}=\{(o_t, a_t)\}_t$, the goal of behavior cloning is to learn a mapping $\pi$ from observation space $\mathcal{O}$ to the action space $\mathcal{A}$.
This map is often learned in a supervised fashion with $\pi$ as a deep neural network minimizing some loss function $\mathcal{L}(\pi(o), a)$ on the observed behavior data pairs $(o, a) \in \mathcal{D}$.
Traditionally, $\mathcal{L}$ was simply taken as the MSE loss, but its inability to admit multiple modes of action for an observation led to different loss formulations~\cite{lynch2020learning, florence2022implicit, shafiullah2022behavior, chi2023diffusion}.
Similarly, understanding that the environment may be partially observable led to modeling the distribution $\mathbb{P}(a_t \mid o_{t-h:t})$ rather than $\mathbb{P}(a_t \mid o_{t})$.
Finally, understanding that such behavior datasets are often generated with an explicit or implicit goal, many recent approaches condition on an (implicit or explicit) goal variable $g$ and learn a goal-conditioned behavior $\mathbb{P}(a \mid o, g)$.
Note that such behavior datasets crucially do not contain any ``reward'' information, which makes this setup different from reward-conditioned learning as a form of offline RL.

\subsection{Behavior Transformers}
\label{sec:behavior_transformers}
Behavior transformer (BeT)~\cite{shafiullah2022behavior} and conditional behavior transformer (C-BeT)~\cite{cui2022play} are respectively two unconditional and goal-conditional behavior cloning algorithms built on top of GPT-like transformer architectures.
In their respective settings, they have shown the ability to handle temporal correlations in the dataset, as well as the presence of multiple modes in the behavior.
While GPT~\cite{brown2020language} itself maps from discrete to discrete domains, BeT can handle multi-modal continuous output space by a clever tokenization trick.
Prior to training, BeT learns a k-means based encoder/decoder that can convert continuous actions into one discrete and one continuous component.
Then, by learning a categorical distribution over the discrete component and combining the component mean with a predicted continuous ``offset'' variable, BeT can functionally learn multiple modes of the data while each mode remains continuous.
While the tokenizer allows BeT handle multi-modal actions, the use of k-means means that choosing a good value of $k$ is important for such algorithms.
In particular, if $k$ is too small then multiple modes of action gets delegated to the same bin, and if $k$ is too large one mode gets split up into multiple bins, both of which may result in a suboptimal policy.
Also, when the action has a large number of (potentially correlated) dimensions, for example when performing action chunking~\cite{zhao2023learning}, non-parametric algorithms like k-means may not capture the nuances of the data distribution.
Such shortcomings of the tokenizer used in BeT and C-BeT is one of the major inspirations behind our work.

\subsection{Residual Vector Quantization}
\label{sec:rvq}
In order to tokenize continuous action, we employ Residual Vector Quantization (Residual VQ) \cite{zeghidour2021soundstream} as a discretization bottleneck.
Vector quantization is a quantization technique where continuous values are replaced by a finite number of potentially learned codebook vectors.
This process maps the input $x$ to an embedding vector $z_q$ in the codebook $\{e_1, e_2, \cdots e_k\}$ by the nearest neighbor look-up:
\begin{equation}
    z_q = e_c,\quad \text{where} \; c=\mathrm{argmin}_j||x-e_j||_2.
\end{equation}
Residual VQ is a multi-stage vector quantizer \cite{vasuki2006review} which replaces each embedding of vanilla VQ-VAE \cite{van2017neural} with the sum of vectors from a finite layers of codebooks.
This approach cascades $N_q$ layers of vector quantizations residually: the input vector $x$ is passed through the first stage of vector quantization to derive $z_q^{1}$.
The residual, $x - z_q^{1}$, is then iteratively quantized by a sequence of $N_q-1$ quantizing layers, passing the updated residual $x - \sum_{i=1}^{p} z_q^{i}$ to the next layer.
The final quantized input vector is then the sum of vectors from a set of finite codebooks $z_q(x) = \sum_{i=1}^{N_q} z_q^{i}$.

\section{Vector-Quantized Behavior Transformers}
\begin{figure*}[t]
\begin{center}
\centerline{\includegraphics[width=1.\linewidth]{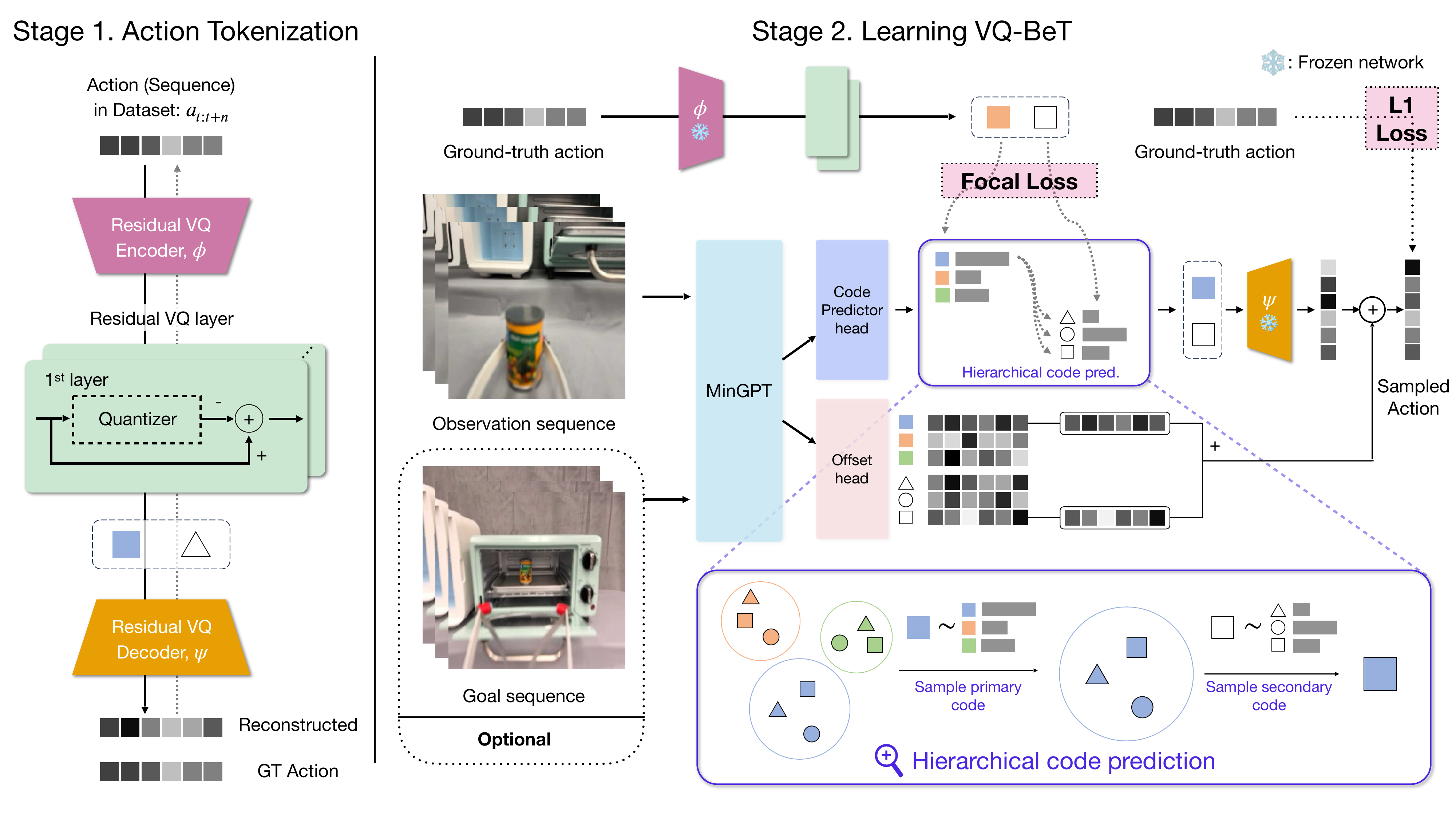}}
\vskip -0.15in
\caption{Overview of~\method{}, broken down into the residual VQ encoder-decoder training phase and the ~\method{} training phase. The same architecture works for both conditional and unconditional cases with an optional goal input. In the bottom right, we show a detailed view of the hierarchical code prediction method.}
\vskip -0.3in
\label{fig2:overview}
\end{center}
\end{figure*}

In this section, we introduce~\method{}, which has capability to solve both conditional and non-conditional tasks from uncurated behavior dataset.~\method{} is composed of two stages: Action discretization phase (stage 1 in Figure \ref{fig2:overview}) and~\method{} learning phase (stage 2 in Figure \ref{fig2:overview}). Each stage is explained in Section \ref{rvq_pretraining} and \ref{weighted_update}, respectively.

\subsection{Sequential prediction on behavior data}

Binning actions to tokenize them and predicting the tokenized class has been successfully applied for learning multi-modal behavior \cite{shafiullah2022behavior, cui2022play}.
However, these k-means binning approaches face issues while scaling, as disucssed in Section~\ref{sec:behavior_transformers}.

As such, we propose instead to learn a discrete latent embedding space for action or action chunks, and modeling such action latents instead.
Note that, such latent models in the form of VQ-VAEs and latent diffusion models are widely used in multiple generative modeling subfields, including image, music, and video~\cite{bar2024lumiere,ziv2024masked,podell2023sdxl}.
With such discrete tokenziation, our model can directly predict action tokens from observation sequences optionally conditioned on goal vectors.

\subsection{Action (chunk) discretization via Residual VQ}\label{rvq_pretraining}

We employ Residual VQ-VAE \cite{zeghidour2021soundstream} to learn a scalable action discretizer and address the complexity of action spaces encountered in the real world.
The quantization process of an action (or action chunk, where $n > 1$) $a_{t:t+n}$ is learned via learning a pair of encoder and decoder networks; $\phi, \psi$. 
We start with passing $a_{t:t+n}$ through the encoder $\phi$.
The resulting latent embedding vector $x=\phi(a_{t:t+n})$ is then mapped to an embedding vector in the codebook of the first layer $z_q^1 \in \{e_1^1, \cdots e_k^1\}$ by the nearest neighbor look-up, and the residual is recursively mapped to each codebook of the remaining $N_q-1$ layers $z_q^i \in \{e_1^i, \cdots e_k^i\}$, where $i=2,\cdots,N_q$.
The latent embedding vector $x=\phi(a_{t:t+n})$ is represented by the sum of vectors from codebooks $z_q(x) = \sum_{i=1}^{N_q} z_q^i$, where each vector $z_q^{i=1:N_q}$ works as the centroid of hierarchical clustering.

Then, the discretized vector $z_q(x)=\sum_{i=1}^{N_q} z_q^i$  is reconstructed as $\psi(z_q(x))$ by passing through the decoder $\psi$. We train Residual VQ-VAE using a loss function, as shown in Eq \ref{eq1:rvq_loss}.
The first term represents the reconstruction loss, and the second term is the VQ objective that shifts the embedding vector $e$ towards the encoded action $x=\phi(a_{t:t+n})$.
To update the embedding vectors $e_{1:k}^{1:N_q}$, we use moving averages rather than direct gradient updates following \cite{islam2022discrete, mazzaglia2022choreographer}.
In all of our experiments, it was sufficient to use $N_q: = 2$ VQ-residual layers, and keep the commitment loss $\lambda_\mathrm{commit}:=1$ constant.
\begin{align}
    \mathcal{L}_\mathrm{Recon} = & \left\lVert a_{t:t+n}-\psi(z_q(\phi(a_{t:t+n})))\right\lVert_1 \\
    \mathcal{L}_\mathrm{RVQ} = & \mathcal{L}_\mathrm{Recon} + \left\|\mathrm{SG}[\phi(a_{t:t+n})]-e\right\|_2^2 \label{eq1:rvq_loss} \\ + \lambda_\mathrm{commit}&\|\phi(a_{t:t+n})-\mathrm{SG}[e]\|_2^2  , \quad \mathrm{(SG : stop\; gradient)}\nonumber \label{eq2:recon_loss}
\end{align}
We indicate the codes of the first quantizer layer as \textit{primary code}, and the codes of the remaining layers as \textit{secondary codes}. 
Intuitively, the primary codes in Residual VQ performs coarse clustering over a large range within the dataset, while the secondary codes handle fine-grained actions. (Decoded centroids are visualized in Appendix Figure~\ref{fig:decoded_centroids}.)

\subsection{Weighted update for code prediction}\label{weighted_update}

After training Residual VQ, we train GPT-like transformer
architecture to model the probability distribution of action or action chunks from the sequence of observations.
One of the main differences between BeT and~\method{} stems from using a learned latent space.
Since our vector quantization codebooks let us freely translate between an action latent $z_q(\phi(a_{t:t+n})) = \sum_{i=1}^{N_q} z_q^i$ and the sequence of chosen codes at each codebook, $\{z_q^{i}\}_{i=1}^{N_q}$, we use 
them as a labels in the code prediction $\mathcal{L}_{\mathrm{code}}$ loss to learn the categorical prediction head $\zeta_{\mathrm{code}}^{i}$ for given sequence of observations $o_{t-h:t}$.
Following \cite{shafiullah2022behavior, cui2022play}, we employ Focal loss \cite{lin2017focal} to train the code prediction head by comparing the probabilities of the predicted categorical distribution with the actual labels $z_q^{i}$.
We adjust the weights between the primary code and secondary code learning losses, leveraging our priors about the latent space.
\begin{equation}\label{eq:code_loss}
    \mathcal{L}_{\mathrm{code}} = \mathcal{L}_{\mathrm{focal}}(\zeta_{\mathrm{code}}^{i=1}(o_{t})) + \beta \mathcal{L}_\mathrm{focal}(\zeta_{\mathrm{code}}^{i>1}(o_{t}))
\end{equation}
Finally, the quantized behavior is obtained by passing the sum of the predicted residual embeddings through the decoder as follows.
\begin{equation}\label{eq:quantized_behavior_representation}
    \lfloor a_{t:t+n} \rfloor = \psi \Big( \sum_{j, i} e_j^i \cdot \mathbb I [\zeta_{\mathrm{code}}^{i}=j)] \Big)
\end{equation}
We adopt additional offset head $\zeta_{\mathrm{offset}}$ to maintain full fidelity, adjusting the centers of discretized actions based on observations. The total~\method{} loss is shown in Eq.~\ref{vqbet_total_loss}.
\begin{equation}\label{eq:offset_loss}
    \mathcal{L}_{\mathrm{offset}} = \Big| a_{t:t+n}-\big(\left\lfloor a_{t:t+n} \right \rfloor + \zeta_\mathrm{offset}\left(o_t\right)\big)\Big|_1
\end{equation}
\begin{equation}
\label{vqbet_total_loss}
    \mathcal{L}_{\mathrm{~\method{}}} = \mathcal{L}_{\mathrm{code}} + \mathcal{L}_{\mathrm{offset}}
\end{equation}
\subsection{Conditional and non-conditional task formulation}
To provide a general-purpose behavior-learning model that can predict multi-modal continuous actions in both conditional and unconditional tasks, we introduce conditional and non-conditional task formulation of ~\method{}.

\begin{figure*}[t!]
\begin{center}
\centerline{\includegraphics[width=1.\linewidth]{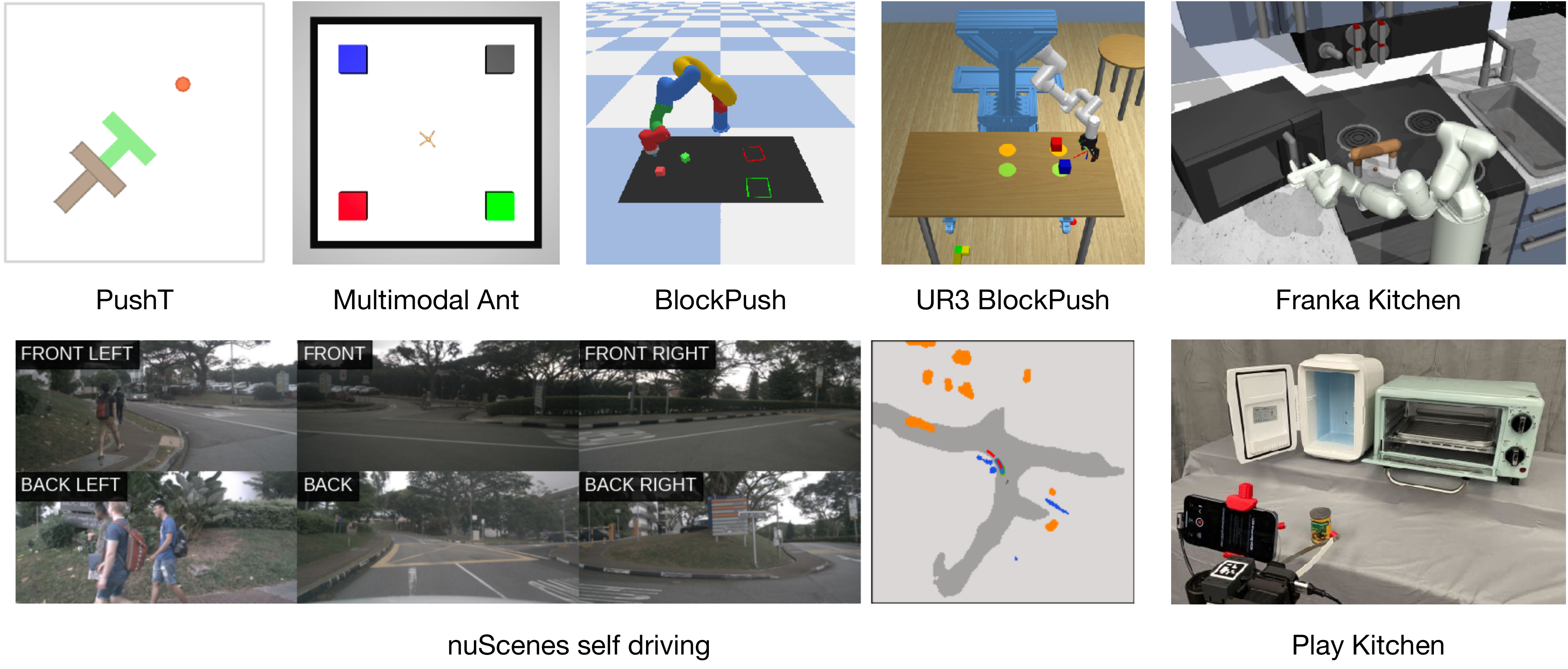}}
\vskip -0.16in
\caption{Visualization of the environments (simulated and real) where we evaluate~\method{}. Top row contains PushT~\cite{chi2023diffusion}, Multimodal Ant~\cite{brockman2016openai}, BlockPush~\cite{florence2022implicit}, UR3 BlockPush~\cite{kim2022automating}, Franka Kitchen~\cite{gupta2019relay}, and bottom row contains nuScenes self-driving~\cite{caesar2020nuscenes}, and our real robot environment.}
\label{fig:environments}
\vspace{-2.5em}
\end{center}
\end{figure*}

\paragraph{Non-conditional formulation:}

For a given dataset $\mathcal{D}=\{o_t, a_t\}$, we consider a problem of predicting the distribution of possible action sequences $a_{t:t+n}$ conditioned on a sampled sequence of observations $o_{t-h:t}$.
Thus, we formulate the behavior policy as $\pi: \mathcal{O}^h \rightarrow \mathcal{A}^n$, where $\mathcal{O}$ and $\mathcal{A}$ denotes the observation space and action space, respectively.
\paragraph{Conditional formulation:}
For goal-conditional tasks, we extend the formulation above to take a goal conditioning vector in the form of one or more observations.
Given current observation sequence and future observation sequence, we now consider an extended policy model that predicts the distribution of sequential behavior $\pi: \mathcal{O}^h \times \mathcal{O}^g \rightarrow \mathcal{A}^n$, where $o_{t-h:t} \in \mathcal{O}^h$ and $o_{N-g:N} \in \mathcal{O}^g$ are current and future observation sequences.

\begin{table}[ht]
\resizebox{\columnwidth}{!}{%
\begin{tabular}{@{}lcccccc@{}}
\toprule
Environment    & Metric                                                                                 & GCBC & C-BeT & C-BESO        & CFG-BESO & VQ-BeT        \\ \midrule
PushT          & \multirow{2}{*}{\begin{tabular}[c]{@{}c@{}}Final IoU\\ ($\cdot/1$)\end{tabular}}       & 0.02 & 0.02  & 0.30           & 0.25     & \textbf{0.39} \\
Image PushT    &                                                                                        & 0.02 & 0.01  & 0.02          & 0.01     & \textbf{0.10}  \\ \midrule
Kitchen        & \multirow{2}{*}{\begin{tabular}[c]{@{}c@{}}Goals\\ ($\cdot/4$)\end{tabular}} & 0.15 & 3.09  & 3.75          & 3.47     & \textbf{3.78} \\
Image Kitchen  &                                                                                        & 0.64 & 2.41  & 2.00             & 1.59     & \textbf{2.60}  \\ \midrule
Multimodal Ant & \multirow{2}{*}{\begin{tabular}[c]{@{}c@{}}Goals\\ ($\cdot/2$)\end{tabular}} & 0.00    & 1.68  & 1.14          & 0.92     & \textbf{1.72} \\
UR3 BlockPush  &                                                                                        & 0.19 & 1.67  & \textbf{1.94} & 1.91     & \textbf{1.94} \\ \midrule
BlockPush      & Success  ($\cdot/1$)                                                              & 0.01 & 0.87  & \textbf{0.93} & 0.88     & 0.87          \\ \bottomrule
\end{tabular}
}
\vspace{-0.5em}
\caption{Comparing different algorithms in goal-conditional behavior generation. The seven simulated robotic manipulation and locomotion environments used here are described in Section~\ref{sec:environments}.}
\label{tab:conditional}
\vspace{-0.5em}
\end{table}

\section{Experiments}
\label{experiments}
With both conditional and unconditional \method{}, we run experiments to understand how well they can model behavior on different datasets and environments.
We focus on two primary properties of ~\method{}'s generated behaviors: quality, as evaluated by how well the generated behavior achieves some task objective or goal, and the diversity, as evaluated by the entropy of the distribution of accomplished subtasks or goals.
Concretely, through our experiments, we try to answer the following questions:
\begin{enumerate}
\itemsep0em 
    \item How well do~\method{} policies perform on the respective environments in both conditional and unconditional behavior generation?
    \item How well does~\method{} capture the multi-modality present in the dataset?
    \item Does~\method{} scale beyond simulated tasks?
    \item What design choices of ~\method{} make the most impact in its performance?
\end{enumerate}

\subsection{Environments, datasets, and baselines}
\label{sec:environments}
\begin{table}[t!]
\resizebox{\columnwidth}{!}{%
\begin{tabular}{@{}lcccccc@{}}
\toprule
Environment   & Metric & BC   & BeT  & DiffPolicy-C  & DiffPolicy-T  & VQ-BeT   \\ \midrule
PushT         & \multirow{2}{*}{\begin{tabular}[c]{@{}c@{}}Final IoU\\ ($\cdot / 1$)\end{tabular}} & 0.65 & 0.39 & 0.73 & 0.74 & \textbf{0.78} \\
Image PushT   &        & 0.13 & 0.01 & 0.66          & 0.45          & \textbf{0.68} \\ \midrule
Kitchen       & \multirow{3}{*}{\begin{tabular}[c]{@{}c@{}}Goals\\ ($\cdot/4$)\end{tabular}}       & 0.18 & 3.07 & 2.62 & 3.44 & \textbf{3.66} \\
Image Kitchen &        & 0.75 & 2.48 & \textbf{3.11} & 3.01          & 2.98          \\
Multimodal Ant           &        & 0.01 & 2.73 & 3.12          & 2.90           & \textbf{3.22} \\ \midrule
UR3 BlockPush & \multirow{2}{*}{\begin{tabular}[c]{@{}c@{}}Goals\\ ($\cdot/2$)\end{tabular}}       & 0.11 & 1.59 & 1.83 & 1.82 & \textbf{1.84} \\
BlockPush     &        & 0.01 & 1.67 & 0.47          & \textbf{1.93} & 1.79          \\ \bottomrule
\end{tabular}%
}
\vspace{-0.5em}
\caption{Performance of different algorithms in unconditional behavior generation tasks. We evaluate over seven simulated robotic manipulation and locomotion tasks as described in Section~\ref{sec:environments}.}
\label{tab:unconditional}
\vspace{-1em}
\end{table}

Across our experiments, we use a variety of environments and datasets to evaluate~\method{} (Figure \ref{fig:environments}). In simulation, we evaluate the wider applicability of VQ-BeT on eight benchmarks; namely, six manipulation tasks including two image-based tasks: (a) PushT, (b) Image PushT, (c) Kitchen, (d) Image Kitchen, (e) UR3 BlockPush, (f) BlockPush; a locomotion task,~(g) Multimodal Ant; and a self-driving benchmark, (h) NuScenes. The environments are visualized in Figure~\ref{fig:environments}, and a detailed descriptions of each task is provided in Appendix~\ref{Simulated environments}. We also evaluate on a real-world environment with twelve tasks (five single-phase, three multi-phase tasks and four long-horizon tasks) described in Section~\ref{sec:real-robots}.

\paragraph{Baselines:} We compare \method{} against the SOTA methods in behavior modeling in both conditional and unconditional categories. In both of these categories, we compare against transformer- and diffusion-based baselines.

For unconditional behavior generation, we compare against MLP-based behavior cloning, the original Behavior Transformers (BeT)~\cite{shafiullah2022behavior} and Diffusion Policy~\cite{chi2023diffusion}.
The BeT architecture uses a k-means tokenization as explained in Section~\ref{sec:behavior_transformers}.
Diffusion policy~\cite{chi2023diffusion}, on the other hand, uses a denoising diffusion head~\cite{ho2020denoising} to model multi-modality in the behaviors. We use both the convolutional and transformer variant of the diffusion policy as baselines for our work since they excel in different cases.

For {goal-conditional behaviors,} we compare against simple goal conditioned BC, Conditional Behavior Transformers (C-BeT)~\cite{cui2022play} and BESO~\cite{reuss2023goal}. C-BeT uses k-means tokenization but otherwise has a similar architecture to ours. BESO uses denoising diffusion, and has a conditioned variant (C-BESO) and a classifier-free guided variant (CFG-BESO) that we compare against.

\subsection{Performance of behavior generated by \method{}}

We evaluate \method{} in a set of goal-conditional tasks in Table~\ref{tab:conditional} and a set of unconditional tasks in Table~\ref{tab:unconditional}.
On the PushT environments, we look at final and max coverage, where the coverage value is the IoU between the T block and the target T position.
For the unconditional Kitchen, BlockPush, and Ant tasks, we look at the total number of tasks completed in expectation, where the maximum possible number of tasks is 4, 2, and 4 respectively.
For the conditional environments, we report the expected number of successes given a commanded goal sequence, where the numbers of commanded goals are 4 in Kitchen, 2 in Ant, and 2 in BlockPush.
Across all of these metrics, a higher number designates a better performance.

From Tables~\ref{tab:conditional} and~\ref{tab:unconditional}, we see that in both conditional and unconditional tasks, \method{} largely outperforms or matches the baselines.
First, on the conditional tasks, we find that \method{} outperforms all baselines in all tasks except for BlockPush. 
In BlockPush, \method{} performs on par with BeT, while C-BESO and CFG-BESO performs slighly better.
Note that BlockPush has one of the simplest action spaces (2-D $\Delta x, \Delta y$) in the dataset while also having the largest demonstration dataset, and thus the added advantage of having vector quantized actions may not have such a strong edge.
Next, in unconditional tasks, we find that \method{} outperforms all baselines in Franka Kitchen (state), Ant Multimodal, UR3 Multimodal, and both PushT (state and image) environments.
In BlockPush environment, \method{} is outperformed by DiffusionPolicy-T, while in Image Kitchen it is outperformed by DiffusionPolicy-C.
However, \method{} empirically shows stable performances on all tasks, while DiffusionPolicy-T struggles in Image PushT environments, and DiffusionPolicy-C underperforms in Kitchen and BlockPush environments.

\begin{figure}[t!]
\begin{center}
\centerline{\includegraphics[width=1.\linewidth]{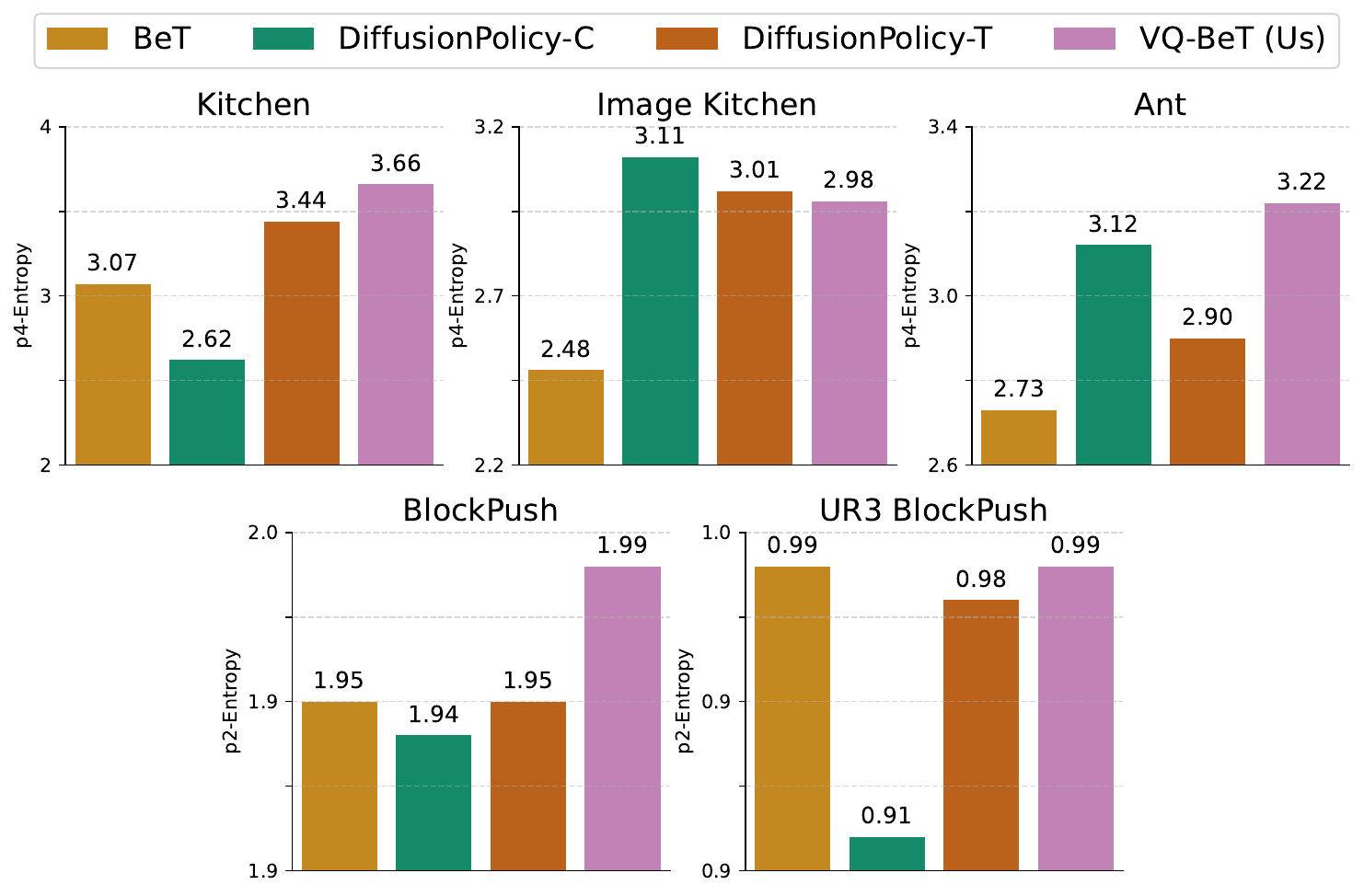}}
\vskip -0.1in
\caption{A comparison between the behavior entropy of the algorithms, calculated based on their task completion order, on five of our simulated environments.}
\vskip -0.3in
\end{center}
\end{figure}

\subsection{How well does \method{} capture multimodality?}
One of the primary promises of behavior generation models is to capture the diversity present in the data, rather than simply copying a single mode of the existing data very well.
Thus, for a quantitative measure we examine the behavior entropy of the models in the unconditional behavior generation task.
Behavior entropy here tries to captures the diversity of a model's generated long horizon behaviors.
We compare the final-subtask entropy as a balanced metric between performance and diversity.
We see that~\method{} outperforms all baselines in all tasks except for Image Kitchen, where it's outperformed by DiffusionPolicy-T.
However, behavior diversity is hard to capture properly in a single number, which is why we also present the diversity of generated behavior on the PushT task in Figure~\ref{fig:intro} (left).
There, we can see how~\method{} captures both modes of the dataset in rollouts, while also generating overall smooth trajectories.

\subsection{Inference-time efficiency of \method{}}
\begin{table}[h]
\vskip -0.06in
\resizebox{\columnwidth}{!}{%
\footnotesize
\begin{tabular}{@{}lcccc@{}}
\toprule
Unconditional & C-BeT  & C-BESO    & CFG-BESO  &~\method{} \\ \midrule
Single step   & 22.6ms & 25.9ms    & 41.7ms    & 22.8ms \\
Multi step    & \ding{55}      & \ding{55}         & \ding{55}         & 23.3ms \\ \midrule
Conditional   & BeT    & DiffusionPolicy-C & DiffusionPolicy-T &~\method{} \\ \midrule
Single step   & 13.2ms & 100.5ms   &  98.6ms   & 15.1ms \\
Multi step    & \ding{55}      & 100.7ms    & 98.6ms    & 15.2ms \\ %
\end{tabular}
}
\caption{Inference times for~\method{} and baselines in kitchen environment. For DiffusionPolicy we rolled-out with 10-iteration diffusion, following their real-world settings. The methods that only support single-step action prediction are marked with \ding{55}.}
\label{tab:infertime}
\end{table}

Denoising diffusion based models such as DiffusionPolicy and BESO require multiple forward passes from the network to generate a single action or action chunk.
In contrast,~\method{} can generate action or action chunks in a single forward pass.
As a result,~\method{} enjoys much faster inference times, as shown in Table~\ref{tab:infertime}.
Receding horizon control using action chunking can speed up some of our baselines, but~\method{} can take advantage of the same, speeding up the method proportionally.
Moreover, receding horizon control is not a silver bullet; it can be problematic in affordable, inaccurate hardware, as we show in Section~\ref{sec:real-robots} in our real world experiments.

\begin{table}
\resizebox{\columnwidth}{!}{%
\begin{tabular}{@{}lccc@{}}
Method          & \begin{tabular}[c]{@{}c@{}}Access to\\information\end{tabular}              & \begin{tabular}[c]{@{}c@{}}Avg. $L_2$\\ (m) (↓)\end{tabular} & \begin{tabular}[c]{@{}c@{}}Avg. collision\\ (\%) (↓)\end{tabular} \\ \midrule
FF ~\cite{hu2021safe}             & \multirow{4}{*}{Full}                         & 1.43        & 0.43                    \\
EO~\cite{khurana2022differentiable}              &                          & 1.6         & 0.33                    \\
UniAD~\cite{hu2023planning}          &                          & 1.03        & 0.31                    \\
Agent-Driver~\cite{mao2023language}    &                          & \uline{0.74}        & \uline{0.21}                    \\ \midrule
GPT-Driver~\cite{mao2023gpt}      & \multirow{3}{*}{Partial} & 0.84        & 0.44                    \\
Diffusion-based traj. model&                          & 0.96        & 0.49                    \\
VQ-BeT          &                          & \textbf{0.73}        & \textbf{0.29}                    \\ \bottomrule
\end{tabular}%
}
\vspace{-0.06in}
\caption{\textit{(Lower is better)} Trajectory planning performance on the nuScenes environment. We \textbf{bold} the best partial-information model and \uline{underline} the best full-information model. Even with partial information about the environment, ~\method{} can match or beat the SOTA models on the $L_2$ error metric.}
\label{tab:nuscenes}
\vspace{-0.16in}
\end{table}

\subsection{Adapting \method{} for autonomous driving}
While our previous experiments showed robotic manipulation or locomotion results, learning from multi-modal behavior datasets has wider applications.
We evaluate~\method{} in one such case, in a self-driving trajectory planning task using the nuScenes~\cite{caesar2020nuscenes} dataset. 
In this task, given a few frames of observations, the model must predict the next six frames of an car's location.
While nuScenes usually require the trajectory be predicted from the raw images, we adapted the GPT-Driver~\cite{mao2023gpt} framework which uses pretrained models to extract vehicle and obstacle locations and velocities. 
However, this processing also discards road lane and shoulder informations, which makes collision avoidance hard.

In Table~\ref{tab:nuscenes}, we show the performance of ~\method{} in this task, measured by how closely it followed the ground truth trajectory in test scenes, as well as how likely the generated trajectory was to collide with the environment. Note that collision avoidance is especially difficult for agents with partial information since they do not have any lane information.
We find that~\method{} outperforms all other methods in trajectory following, achieving the lowest average $L_2$ distance between the ground truth trajectories and generated trajectories.
Moreover, ~\method{} achieves a collision probability that is better or on-par with older self-driving methods, while not being designed for self-driving in particular.

\subsection{Design decisions that matter for~\method{}}
\label{sec:ablation} 
\begin{figure}[t!]
\begin{center}
\centerline{\includegraphics[width=1.\linewidth]{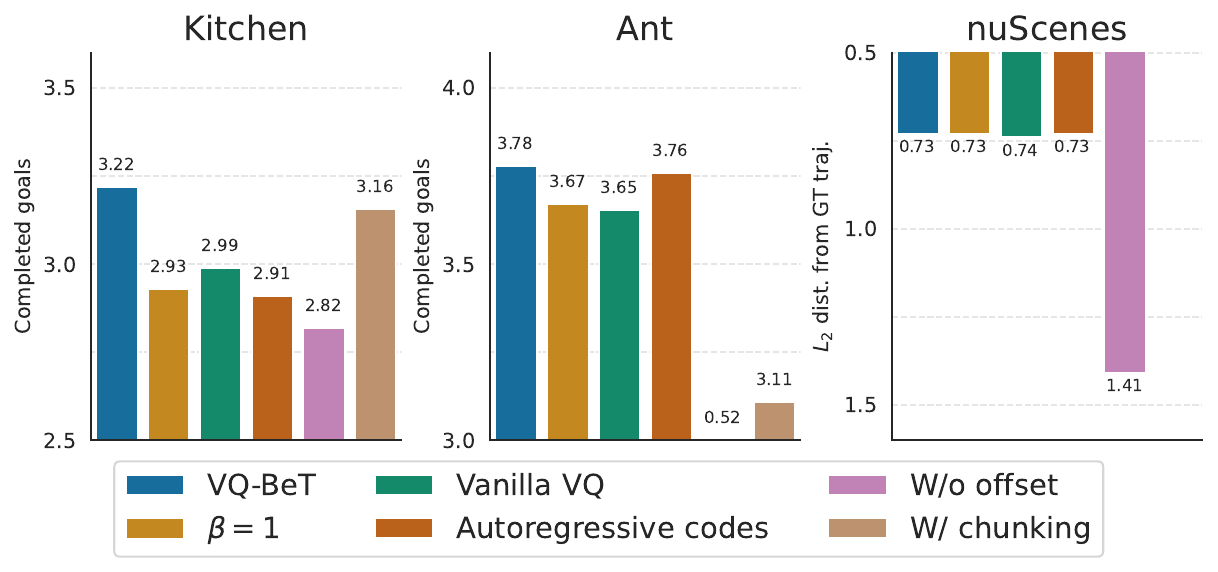}}
\vspace{-1em}
\caption{Summary of our ablation experiments. The five different axes of ablation is described in Section~\ref{sec:ablation}.}
\label{fig:ablation}
\vspace{-2.8em}
\end{center}
\end{figure}

In this section, we examine how changes in each module of VQ-BeT affect its performance. We ablate the following components: using residual vs. vanilla VQ, using an offset head, using action chunking, predicting the VQ-codes autoregressively, and weighing primary and secondary codes equally by setting $\beta = 1$ in Eq.~\ref{eq:code_loss}. We perform these ablation experiments in the conditional Kitchen, unconditional Ant, and the nuScenes self-driving task, and the result summary is presented in Figure~\ref{fig:ablation}.

We note that performance-wise, not using a residual VQ layer has a significant negative impact, which we believe is because of the lack of expressivity from a single VQ-layer.
A similar drop in performance shows up when we weigh the two VQ layers equally by setting $\beta = 1$, in Eq.~\ref{eq:code_loss}.
Both experiments seems to provide evidence that important expressivity is conferred on~\method{} using residual VQs.
Next, we note that predicting the VQ-codes autoregressively has a negative impact on the kitchen environment.
This performance drop is anomalous, since in the real world, we found that the autoregressive (and thus causal) prediction of primary and secondary codes is important for good performance.
In the environments where it is possible, we also tried action chunking~\cite{zhao2023learning}; however the performance for such models were lacking.
Since~\method{} models are small and fast, action chunking isn't necessary even when running it on a real robot in real time.
Finally, we found that the offset prediction is quite important for~\method{}, which points to how important full action fidelity is for sequential decision making tasks that we evaluate on.

\subsection{Adapting \method{} to real-world robots}
\label{sec:real-robots}
While our previous experiments evaluated~\method{} in simulated environments, one of the primary potential applications of it is in learning robot policies from human demonstrations.
In this section, we set up a real robot environment, collect some data, and evaluate policies learned using~\method{}. %

\begin{figure*}[t!]
\begin{center}
\centerline{\includegraphics[width=1.\linewidth]{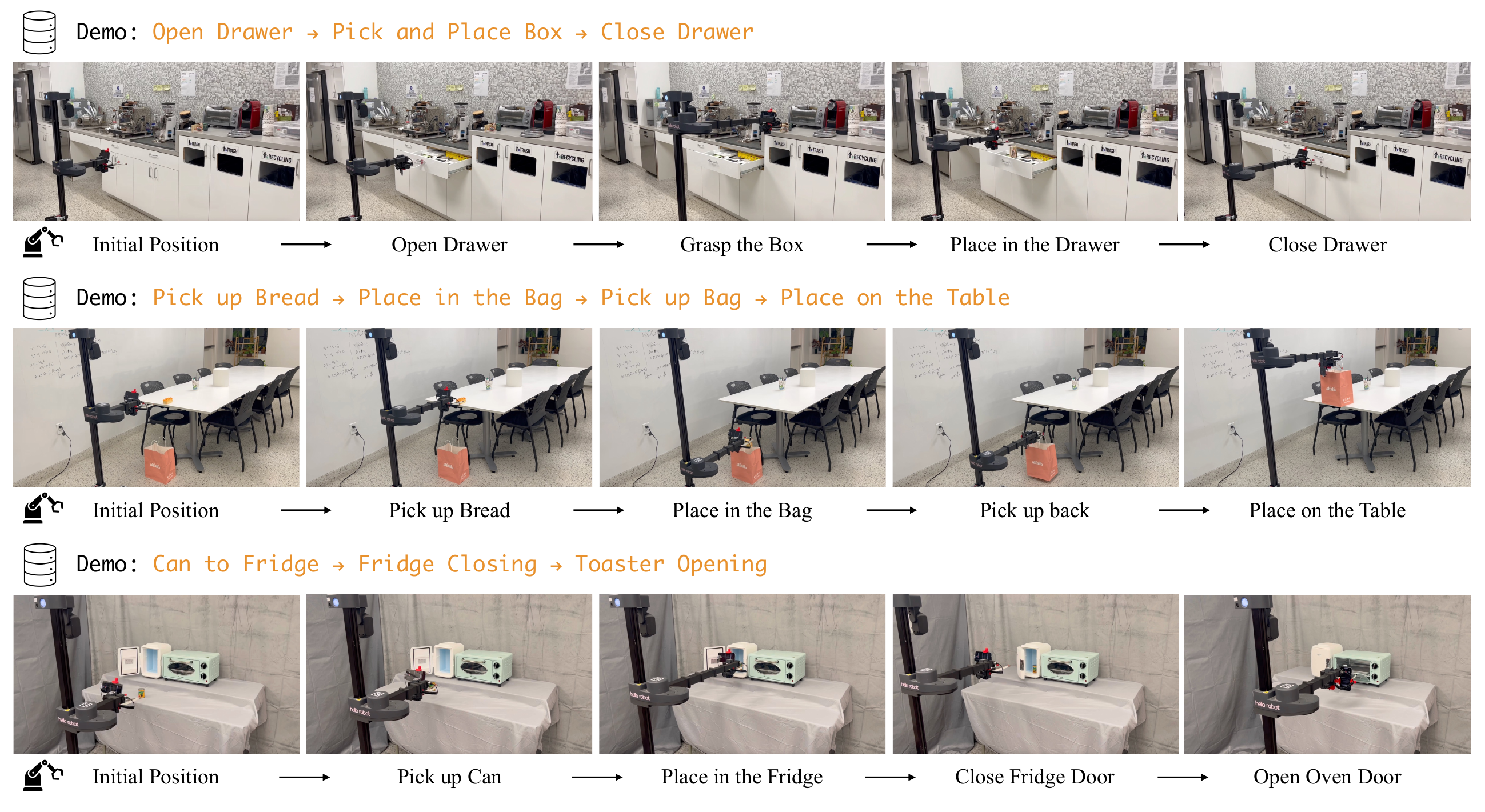}}
\vskip -0.16in
\caption{Visualization of the trajectory VQ-BET generated in a long-horizon real world environment. Each demo consists of three to four consecutive tasks. Please refer to Table \ref{tab:realrobot_long_hor} for the success rates for each task.}
\label{fig:long_hor_environments}
\vspace{-2.5em}
\end{center}
\end{figure*}

\paragraph{Environment and dataset:}

For single-phase and two-phase tasks, we run our experiments in a kitchen-like environment with a toaster oven, a mini-fridge, and a small can in front of the robot as shown in Figure~\ref{fig:environments}. For long-horizon scenarios consisting of more than three tasks, we also test on a real kitchen environment as shown in Figure \ref{fig:long_hor_environments}. We use a similar robot and data collection setup as Dobb·E~\cite{shafiullah2023bringing}, and use the Hello Robot: Stretch~\cite{kemp2022design} for policy rollouts. We create a set of single-phase and multi-phase tasks on this environment (See Table~\ref{tab:realrobot}, or Appendix~\ref{Appendix:Real-world environments} for details). While the single-phase tasks can only be completed in one way, some multi-phase tasks have multi-modal solutions in the benchmark and the datasets.

\paragraph{Baselines:} In this environment, we use MLP-BC and BC with Depth as our simple baselines, and DiffusionPolicy-T as our multi-modal baseline. To handle visual inputs, all models are prepended with the HPR encoder from~\citet{shafiullah2023bringing} which is then fine-tuned during training.

\begin{table}[ht!]
    \centering
    \resizebox{\columnwidth}{!}{%
    \begin{tabular}{@{}lcccccc@{}}
    Method      & Open Toaster  & Close Toaster  & Close Fridge   & Can to Toaster & Can to Fridge & Total \\ \midrule
    VQ-BeT      & $\mathbf{8/10}$ & $\mathbf{10/10}$ & $\mathbf{10/10}$ & $\mathbf{10/10}$ & $9/10$ & $\mathbf{47/50}$          \\
    DiffPol-T$^\dagger$   & $\mathbf{8/10}$          & $9/10$           & $8/10$           & $\mathbf{10/10}$ & $\mathbf{10/10}$ & $45/50$ \\
    BC w/ Depth & $0/10$          & $7/10$           & $\mathbf{10/10}$ & $8/10$           & $2/10$  & $27/50$         \\
    BC          & $0/10$          & $8/10$           & $7/10$           & $9/10$           & $5/10$  & $29/50$         \\ \bottomrule
    \end{tabular}%
    }
    \resizebox{\columnwidth}{!}{%
    \begin{tabular}{@{}lcccc@{}}
    & & & & \\
    Method &
      \begin{tabular}[c]{@{}c@{}}Can to Fridge → \\ Close Fridge\end{tabular} &
      \begin{tabular}[c]{@{}c@{}}Can to Toaster → \\ Close Toaster\end{tabular} &
      \begin{tabular}[c]{@{}c@{}}Close Fridge\\ and Toaster\end{tabular} &
      Total \\ \midrule
    VQ-BeT     & $\mathbf{6/10}  $ & $\mathbf{8/10}  $ & $5/10           $ & $\mathbf{19/30}$ \\
    DiffPol-T$^\dagger$  & $4/10           $ & $1/10           $ & $\mathbf{6/10}  $ & $11/30$ \\
    BC w/ Depth   & $2/10           $ & $0/10           $ & $2/10           $ & $4/30$  \\
    BC         & $2/10           $ & $1/10           $ & $4/10           $ & $7/30$  \\ \bottomrule
    \end{tabular}%
    }
    \vspace{-0.5em}
    \caption{Real world robot experiments solving a number of standalone tasks (top) and two-task sequences (bottom). Here, $\dagger$ denotes that we modified DiffusionPolicy-T to improve its performance; see Section~\ref{sec:real-robots} paragraph ``Practical concerns''.}
    \label{tab:realrobot}
    \vspace{-1em}
\end{table}

\paragraph{Results:} We present the experiment results from the real world environment in Table~\ref{tab:realrobot} and Table~\ref{tab:realrobot_long_hor}. Table~\ref{tab:realrobot} is split in two halves for single-phase and two-phase tasks.
On the single-phase tasks, we see that, simple MLP-BC models are able to perform almost all tasks with some success, which shows that the subtasks are achievable, and the baselines are implemented well.
On these single-phase tasks, \method{}~ marginally outperforms DiffusionPolicy-T, while both algorithms achieve a $\geq 90\%$ success rate.
However, the more interesting comparison is in the two-phase, longer horizon tasks.
Here,~\method{} outperforms all baselines, including DiffusionPolicy, by a relative margin of $73\%$.

Besides comparisons with baselines, we also notice multimodality in the behavior of~\method{}. Especially in the task ``Close Fridge and Toaster'', we note that our model closes the doors in both possible orders during rollouts rather than collapsing to a single mode of behavior.

\begin{table*}[ht!]
    \resizebox{\linewidth}{!}{%
    \centering
    \begin{tabular}{@{}l|cccccccccc@{}}
    \toprule
    Task 1 & Approach Handle  & Grasp Handle  & Open Drawer   & Let Handle Go & Approach the Box & Grasp the Box & Move to Drawer & Place Box inside & Go in front of Drawer & Close Drawer \\ \midrule
    VQ-BeT      & ${8/10}$ & ${7/10}$ & ${7/10}$ & ${7/10}$ & ${7/10}$ & ${7/10}$ & ${7/10}$ & ${6/10}$ & ${6/10}$ & ${6/10}$\\
    DiffPol-T$^\dagger$   & ${10/10}$ & ${9/10}$ & ${9/10}$ & ${9/10}$ & ${8/10}$ & ${3/10}$ & ${3/10}$ & ${3/10}$ & ${3/10}$ & ${2/10}$\\ 
    \bottomrule

    Task 2 & Approach Bread  & Grasp the Bread  & Move to the Bag   & Place Bread inside & Approach the Handle & Grasp the Handle & Lift Bag up & Place on the table & Let Handle go &  \\ \midrule
    VQ-BeT      & ${10/10}$ & ${10/10}$ & ${10/10}$ & ${4/10}$ & ${3/10}$ & ${3/10}$ & ${3/10}$ & ${3/10}$ & ${3/10}$ & \\
    DiffPol-T$^\dagger$   & ${9/10}$ & ${9/10}$ & ${9/10}$ & ${9/10}$ & ${2/10}$ & ${2/10}$ & ${2/10}$ & ${1/10}$ & ${1/10}$ & \\ 
    \bottomrule

    Task 3      & Grasp Can & Pick up Can & Can into Fridge & Let Go of Can & Move Left of Fridge Door & Close Fridge Door & Go in Front of Toaster & Grasp Toaster Handle & Open Toaster & Return to Home Pos. \\ \midrule
    VQ-BeT      & ${10/10}$ & ${10/10}$ & ${10/10}$ & ${8/10}$ & ${8/10}$ & ${8/10}$ & ${8/10}$ & ${7/10}$ & ${7/10}$ & ${7/10}$\\
    DiffPol-T$^\dagger$   & ${5/10}$ & ${5/10}$ & ${5/10}$ & ${4/10}$ & ${2/10}$ & ${2/10}$ & ${2/10}$ & ${2/10}$ & ${2/10}$ & ${2/10}$ \\ 
    \bottomrule

    Task 4      & Grasp Can & Pick up Can & Can into Toaster & Drops Can on Tray & Goes Below Toaster Door & Close Toaster Door & Backs up & Move Left of Fridge Door & Close Fridge & Return to Home Pos. \\ \midrule
    VQ-BeT      & ${10/10}$ & ${10/10}$ & ${8/10}$ & ${8/10}$ & ${8/10}$ & ${6/10}$ & ${6/10}$ & ${6/10}$ & ${6/10}$ & ${6/10}$ \\
    DiffPol-T$^\dagger$   & ${9/10}$ & ${9/10}$ & ${8/10}$ & ${8/10}$ & ${8/10}$ & ${1/10}$ & ${2/10}$ & ${2/10}$ & ${2/10}$ & ${1/10}$\\ 
    \bottomrule
    
    \end{tabular}%
    }

    \vspace{-0.5em}
    \caption{Long-horizon real world robot experiments (Figure \ref{fig:long_hor_environments}). Each task consists of three to four sequences; Task 1 (Open Drawer → Pick and Place Box → Close Drawer), Task 2 (Pick up Bread → Place in the Bag→ Pick up Bag → Place on the Table), Task 3 (Can to Fridge → Fridge Closing → Toaster Opening), and Task 4 (Can to Toaster → Toaster Closing → Fridge Closing). Here, $\dagger$ denotes that we modified DiffusionPolicy-T to improve its performance as explained in Section~\ref{sec:real-robots} paragraph ``Practical concerns''.}
    \label{tab:realrobot_long_hor}
    \vspace{-0.5em}
    
\end{table*}

\begin{table}[ht!]
    \centering
    \footnotesize
    \resizebox{\columnwidth}{!}{%
    $\begin{array}{lcc}
    \hline & \text{RTX A4000 GPU} & \text{4-Core Intel CPU}\\
    \hline \text { VQ-BeT } & 18.06 & 207.25 \\
    \text { DiffusionPolicy-T } & 573.49 & 5243.82 \\
    \text { BC w/ Depth } & 5.66 & 87.28\\
    \text { BC } & 4.73 & 83.28\\
    \hline
    \end{array}$
    }
    \vspace{-0.5em}
    \caption{Average inference time for real robot (in milliseconds). The GPU column is calculated on our workstation while the CPU column is calculated on the Hello Robot's onboard computer.}
    \label{tab:infertime-real}
    \vspace{-1em}
\end{table}

Additionally, we present results from long-horizon real world experiments consisting of a sequence of three or more subtasks in Figure~\ref{fig:long_hor_environments} and Table~\ref{tab:realrobot_long_hor}. We consider interactions with a wider variety of environments (communal kitchen and conference room) and objects (bread, box, bag, and drawer) compared to the single- or two-phase tasks in order to evaluate VQ-BeT in more general scenes. Overall, we see that VQ-BeT has at least thrice the success rate of DiffusionPolicy at the end of all four tasks. For Task 1 and 2, we observe that VQ-BeT gains a performance advantage toward the end of the episode, although VQ-BeT and DiffusionPolicy perform similarly at the beginning of the episodes. Also note that Task 2 is difficult in our ego-only camera setup, since the bag is out of the view while grabbing the bread. For Tasks 3 and 4, we observe that VQ-BeT outperforms DiffusionPolicy in all subtasks and notably, the performance difference is even more pronounced toward the end of the episode. These long-horizon task results continue to suggest that VQ-BeT may overfit less and learn more robust behavior policies in longer horizons tasks.

\paragraph{Practical concerns:} In practice, we noticed that receding-horizon control as used by~\citet{chi2023diffusion} fails completely in our environment (See Appendix Table~\ref{tab:no-receding-control} for comparison to closed loop control).
Our low-cost mobile manipulator robot lacks precise motion control unlike more expensive robot arms like Franka Panda.
This controller noise causes models to go out of distribution during even a short period (three timesteps) of open-loop rollout.
To resolve this, we rolled out every policy fully closed-loop, which resulted in a much larger inference time gap ($25\times$) between \method{} and Diffusion Policy as presented in Table~\ref{tab:infertime-real}.

\section{Related Works}
\label{relatedwork}

\paragraph{Deep generative models for modeling behavior:}
\method{} builds on a long line of works that leveraged tools from generative modeling to learn diverse behaviors.
The earliest examples are in inverse RL literature~\cite{kalakrishnan2013learning,wulfmeier2015maximum,finn2016guided,ho2016generative}, where such tools were used to learn a reward function given example behavior.
Using generative priors for action generationi, such as GMM by ~\citet{lynch2020learning} or EBMs by~\citet{florence2022implicit}, or simply fitting multi-modal action distributions~\cite{singh2020parrot, pertsch2021spirl} became more common with large, human collected behavior datasets~\cite{mandlekar2018roboturk, gupta2019relay}.
Subsequently, a large body of work~\cite{shafiullah2022behavior,cui2022play,pearce2023imitating,chi2023diffusion,reuss2023goal,chen2023playfusion} used generative modeling tools for generalized behavior learning from multi-modal datasets. 
\vspace{-0.1in}
\paragraph{Action reparametrization:}
While~\citet{shafiullah2022behavior} is the closest analogue to~\method{}, the practice of reparametrizing actions for easier or better control goes back to ``bang-bang'' controllers~\cite{bushaw1952differential, bellman1956bang} replacing continuous actions with extreme discrete values. Discretizing each action dimension separately, however, may exponentially explode the action space, which is generally addressed by assuming each action dimension as independent~\cite{tavakoli2018action} or causally dependent~\cite{metz2017discrete}. Without priors on the action space, each of these assumptions may be limiting, which is why later work opted to learn the reparametrization~\cite{singh2020parrot, dadashi2021continuous, luo2023action} similar to~\method{}.
On another hand, options~\cite{sutton1999between, stolle2002learning} abstract actions temporally but can be challenging to learn from data. 
Many applications instead hand-craft primitives as a parametrized action space~\cite{hausknecht2015deep} which may not scale well for different tasks.

\section{Conclusion}
\label{conclusion}
In this work, we introduce~\method{}, a model for learning behavior from open-ended, multi-modal data by tokenizing the action space using a residual VQ-VAE, and then using a transformer model to predict the action tokens.
While we show that~\method{} performs well on a plethora of manipulation, locomotion, and self-driving tasks, an exciting application of such models would be in scaling them up to large behavior datasets containing orders of magnitude more data, environments, and behavior modes.
Finding a shared latent space of actions between different embodiments may let us ``translate'' policies between different robots or even from human to robots.
Finally, a learned, discrete action space may also make real-world RL application faster, which we would like to explore in the future.

\section*{Acknowledgements}
NYU authors are supported by grants from Amazon, Honda, and ONR award numbers N00014-21-1-2404 and N00014-21-1-2758. This work was partly supported by Institute of Information \& communications Technology Planning \& Evaluation (IITP) grant funded by the Korea government(MSIT) [NO.2021-0-01343, Artificial Intelligence Graduate School Program (Seoul National University)]. NMS is supported by the Apple Scholar in AI/ML Fellowship. LP is supported by the Packard Fellowship.
We thank Jonghae Park for his help in obtaining the UR3 Multimodal dataset.

\section*{Impact Statement}
This paper presents work whose goal is to advance the field of Machine Learning. There are many potential societal consequences of our work, none which we feel must be specifically highlighted here.

\bibliography{main}
\bibliographystyle{icml2024}

\newpage

\appendix
\onecolumn

\newpage
\appendix

\section{Experimental and Dataset}
\subsection{Simulated environments}\label{Simulated environments}
Across our experiments, we use a variety of environments and datasets to evaluate~\method{}.
We give a short descriptions of them here, and depiction of them in Figure~\ref{fig:environments}:
\begin{itemize}[leftmargin=12pt]
\itemsep0em 
    \item \textbf{Franka Kitchen:} We use the Franka Kitchen robotic manipulation environment introduced in~\cite{gupta2019relay} with a Franka Panda arm with a 7 dimensional action space and 566 human collected demonstrations.
    This environment has seven possible tasks, and each trajectory completes a collection of four tasks in some order.
    While the original environment is state-based, we create an image-based variant of it by rendering the states with the MuJoCo renderer as an 112 by 112 image.
    In the conditional variant of the environment, the model is conditioned with future states or image goals (Image Kitchen).
    \item \textbf{PushT:} We adopt the PushT environment introduced in~\cite{chi2023diffusion} where the goal is to push a T-shaped block on a table to a target position. The action space here is two-dimensional end-effector velocity control. Similar to the previous environment, we create an image based variant of the environment by rendering it, and a goal conditioned variant of the environment by conditioning the model with a final position. This dataset has 206 demonstrations collected by humans.
    \item \textbf{BlockPush:} The BlockPush environment was introduced by~\citet{florence2022implicit} where the goal of the robot is to push two red and green blocks into two (red and green) target squares in either order.
    The conditional variant is conditioned by the target positions of the two blocks.
    The training dataset here consists of 1,000 trajectories, with an equal split between all four possibilities of (block target, push order) combinations, collected by a pre-programmed primitive.
    \item \textbf{UR3 BlockPush:} In this task, an UR3 robot tries to move two blocks to two goal circles on the other side of the table \cite{kim2022automating}. Each demonstration is multimodality, since either block can move first. In the non-conditional setting, we evaluate whether each block reaches the goal, while in the conditional setting, we evaluate in which order the blocks get to the given target point.
    \item \textbf{Multimodal Ant:} We adopt a locomotion task that requires the MuJoCo Ant~\cite{brockman2016openai} robot to reach goals located at each corner of the map. The demonstration contains trajectories that reach the four goals in different orders. In the conditional setting, the performance is evaluated by reaching two goals given by the environment, while in the unconditional setting, the agent tries to reach all four goals.
    \item \textbf{nuScenes self-driving:} Finally, to evaluate~\method{} on environments beyond robotics, we use the nuScenes~\cite{caesar2020nuscenes} self-driving environment as a test setup. We use the preprocessed, object-centric dataset from~\citet{mao2023gpt} with 684 demonstration scenes where the policy must predict the next six timesteps of the driving trajectory. In this environment, the trajectories are all goal-directed, where the goal of which direction to drive is given to the policy at rollout time. In Appendix Section~\ref{app:sec:driving-howto}, we detail how we process the GPT-Driver~\citet{mao2023gpt} dataset for use in our method.
\end{itemize}

\subsection{Real-world environments}\label{Appendix:Real-world environments}
We run our experiments on a kitchen-like environment, with a toaster oven, a mini-fridge, and a small can in front of them, as seen in Fig.~\ref{fig:environments}.
In this environment, we define the tasks as opening or closing the fridge or toaster, and moving the can from the table to the fridge or toaster and vice versa.
During data collection and evaluation, the starting position for the gripper and the position of the cans are randomized within a predefined area, while the location of the fridge and the toaster stays fixed.
We use a similar robot and data collection setup as Dobb·E~\cite{shafiullah2023bringing}, using the Stick to collect 45 demonstrations for each task, using 80\% of them for training and 20\% for validation, and using the Hello Robot: Stretch~\cite{kemp2022design} for policy rollouts.
While some of the single tasks can only be completed in one way, the we also test the model on sequences of two tasks, for example closing oven and fridge, which can be completed in multiple ways.
This task multi-modality is also captured in the dataset: tasks that can be completed in multiple ways have multi-modal demonstration data.
\newpage
\section{Additional Results}
\begin{table*}[h!]
\centering
\small
\begin{tabular}
{c|c|ccccccc|c}
\noalign{\smallskip}\noalign{\smallskip}
\multicolumn{2}{c|}{} & C-BeT & C-BESO & CFG-BESO & \textbf{\method{}} \\
\Xhline{2\arrayrulewidth}
\multirow{3}{*}{Kitchen} 
& Full & 3.09 & {3.75}  & {3.47} & \textbf{3.78}    \\
& 1/4 & 2.77 & 2.62 & {3.07} & \textbf{3.46} \\
& 1/10 & 2.59 & 2.67 & {2.73} & \textbf{2.95} \\
\hline
\multirow{1}{*}{Image Kitchen} 
& Full & 2.41 & 2.00  & {1.59} & \textbf{2.60}    \\
\hline
\multirow{3}{*}{Ant Multimodal} 
& Full & 1.68 & {1.14} & 0.92 & \textbf{1.72}   \\
& 1/4 & 0.85 & {0.58} & 0.52 & \textbf{1.23} \\
& 1/10 & 0.35 & {0.39} & 0.40 & \textbf{1.06} \\
\hline
\multirow{3}{*}{BlockPush Multimodal} 
& Full & 0.87 & \textbf{0.93}&  0.88 & 0.87  \\
& 1/4 & 0.48 & {0.52} &  0.47 & \textbf{0.62}  \\
& 1/10 & 0.10 & \textbf{0.29} &  0.17 & 0.13  \\
\hline
\multirow{3}{*}{UR3 Multimodal} 
& $-\ell_1$ & -0.129 & -0.090 & -0.091 & \textbf{-0.085} \\
& p1 &  \textbf{1.00} & 0.98 & 0.97 & \textbf{1.00} \\
& p2 &  0.67 & \textbf{0.96} & 0.94 & 0.94 \\
\hline
\multirow{2}{*}{PushT} 
& Final Coverage & 0.02 & 0.30 & 0.25 & \textbf{0.39} \\
& Max Coverage & 0.11 & 0.41 & 0.38 & \textbf{0.49} \\
\hline
\multirow{2}{*}{Image PushT} 
& Final Coverage & 0.01 & 0.02 & 0.01 & \textbf{0.10} \\
& Max Coverage & 0.02 & 0.02 & 0.02 & \textbf{0.12} \\
\Xhline{2\arrayrulewidth}
\end{tabular}
\caption{Quantitative results of VQ-BeT and related baselines on  conditional tasks. }
\label{tab:conditional_table}
\end{table*}

\begin{table*}[h!]
\centering
\small
\begin{tabular}
{c|c|cccccc|c}
\noalign{\smallskip}\noalign{\smallskip}
\Xhline{2\arrayrulewidth}
\multicolumn{2}{c|}{} & BeT & DiffusionPolicy-C & DiffusionPolicy-T & \textbf{VQ-BeT} \\
\Xhline{2\arrayrulewidth}

\multirow{2}{*}{PushT} 
& Final Coverage & 0.39 & 0.73 & 0.74 & \textbf{0.78}   \\
& Max Coverage & 0.73 & \textbf{0.86} & 0.83 & {0.80}   \\
\hline
\multirow{2}{*}{Image PushT} 
& Final Coverage & 0.01 & 0.66 & 0.45 & \textbf{0.68}   \\
& Max Coverage & 0.01 & \textbf{0.82} & 0.71 & {0.73}   \\
\hline

\multirow{6}{*}{Kitchen} 
& p1 & 0.99 & 0.94 & 0.99 & \textbf{1.00} \\
& p2 & 0.93 & 0.86 & \textbf{0.98} & \textbf{0.98} \\
& p3 & 0.71 & 0.56 & 0.87 & \textbf{0.91} \\
& p4 & 0.44 & 0.26 & 0.60 & \textbf{0.77} \\
& p3-Entropy & \textbf{3.44} & 3.18 & 3.38 & {3.42} \\
& p4-Entropy & 4.01 & 3.62 & 3.89 & \textbf{4.07} \\
\hline

\multirow{6}{*}{Image Kitchen} 
& p1 & 0.97 & {0.99} & {0.97} & \textbf{1.00} \\
& p2 & 0.73 & \textbf{0.95} & {0.90} & {0.93} \\
& p3 & 0.51 & 0.73 & \textbf{0.75} & {0.67} \\
& p4 & 0.27 & \textbf{0.44} & {0.39} & {0.38} \\
& p3-Entropy & {3.03} & 2.36 & 3.01 & \textbf{3.20} \\
& p4-Entropy & 2.77 & 2.93 & \textbf{3.55} & {3.32} \\
\hline

\multirow{6}{*}{Ant Multimodal} 
& p1 & 0.91 & \textbf{0.96} & 0.87 & {0.94}   \\
& p2 & 0.79 & 0.81 & 0.78 & \textbf{0.83} \\
& p3 & 0.67 & 0.73 & 0.69 & \textbf{0.75} \\
& p4 & 0.36 & 0.62 & 0.56 & \textbf{0.70} \\
& p3-Entropy & {3.89} & 4.26 & \textbf{4.27} & {4.19} \\
& p4-Entropy & 3.55 & 4.18 & 4.11 & \textbf{4.20} \\

\hline
\multirow{3}{*}{BlockPush Multimodal} 
& p1 & 0.96 & 0.36 & \textbf{0.99} & {0.96}   \\
& p2 & {0.71} & 0.11 & \textbf{0.94} & {0.83} \\
& p2-Entropy & 1.95 & 1.94 & 1.95 & \textbf{1.99} \\
\hline

\multirow{3}{*}{UR3 Multimodal} 
& p1 & 0.84 & \textbf{1.00} & \textbf{1.00} & \textbf{1.00}   \\
& p2 & 0.75 & 0.83 & 0.82 & \textbf{0.84}   \\
& p2-Entropy & \textbf{0.99} & 0.91 & 0.98 & \textbf{0.99}   \\

\Xhline{2\arrayrulewidth}

\end{tabular}
\caption{Quantitative results of VQ-BeT and related baselines on non-conditional tasks. }
\label{tab:unconditional_table}
\end{table*}

\begin{table*}[ht]
\centering
\small
\begin{tabular}
{c|c|cccc|cccc}
\noalign{\smallskip}\noalign{\smallskip}
\Xhline{2\arrayrulewidth}

\multicolumn{2}{c|}{\multirow{2}{*}{}}& \multicolumn{4}{c|}{L2  ($\downarrow$)} & \multicolumn{4}{c}{Collision (\%)  ($\downarrow$)}\\

\multicolumn{2}{c|}{} & 1s & 2s & 3s & Avg. & 1s & 2s & 3s & Avg. \\

\hline

\multirow{6}{*}{\textit{ST-P3 metrics}} & ST-P3 \cite{hu2022st} & 1.33 & 2.11 & 2.90 & 2.11 & 0.23 & 0.62 & 1.27 & 0.71 \\
                      & VAD \cite{jiang2023vad} & 0.17 & 0.34 & \textbf{0.60} & \textbf{0.37} & 0.07 & 0.10 & 0.24 & 0.14\\
                      & GPT-Driver \cite{mao2023gpt} & 0.20 & 0.40 & 0.70 & 0.44 & 0.04 & 0.12 & 0.36 & 0.17\\
                      & Agent-Driver \cite{mao2023language} & \textbf{0.16} & 0.34 & 0.61 & \textbf{0.37} & \textbf{0.02} & \textbf{0.07} & \textbf{0.18} & \textbf{0.09}\\
                      & Diffusion-based Traj. Prediction & 0.21 & 0.43 & 0.80 & 0.48 & \textbf{0.01} & \textbf{0.07} & 0.35 & 0.14 \\
                      & VQ-BeT & 0.17 & \textbf{0.33} & \textbf{0.60} & \textbf{0.37}& \textbf{0.02} & 0.11 & 0.34 & 0.16 \\
                      \hline
\multirow{9}{*}{\textit{UniAD metrics}} & NMP \cite{zeng2019end} & - & - & 2.31 & - & - & - & 1.92 & - \\
                      & SA-NMP \cite{wei2021perceive} & - & - & 2.05 & - & - & - & 1.59 & - \\
                      & FF \cite{hu2021safe} & 0.55 & 1.20 & 2.54 & 1.43 & 0.06 & 0.17 & 1.07 & 0.43\\
                      & EO \cite{khurana2022differentiable} & 0.67 & 1.36 & 2.78 & 1.60 & 0.04 & 0.09 & 0.88 & 0.33\\
                      & UniAD \cite{hu2023planning} & 0.48 & 0.96 & 1.65 & 1.03 & 0.05 & 0.17 & 0.71 & 0.31\\
                      & GPT-Driver \cite{mao2023gpt} & {0.27} & 0.74 & {1.52} & 0.84 & 0.07 &  {0.15} & 1.10 & 0.44\\
                      & Agent-Driver \cite{mao2023language}  & \textbf{0.22} &  {0.65} & \textbf{1.34} &  {0.74} &  {0.02} & \textbf{0.13} & \textbf{0.48} & \textbf{0.21}\\
                      & Diffusion-based Traj. Prediction &  {0.27} & 0.78 & 1.83 & 0.96 & \textbf{0.00} & 0.27 & 1.21 & 0.49 \\
                      & VQ-BeT & \textbf{0.22} & \textbf{0.62} & \textbf{1.34} & \textbf{0.73} &  {0.02} & 0.16 &  {0.70} &  {0.29}\\
                      \hline
\end{tabular}
\caption{\textit{(Lower is better)} Trajectory planning performance on the nuScenes~\cite{caesar2020nuscenes} self-driving environment. We \textbf{bold} the best performing model. Note that while Agent-Driver outperforms us in some Collision avoidance benchmarks, it is because they use a lot more information than what is available to our agent, namely the road lanes and the shoulders information, without which avoiding collision is difficult for our model or GPT-Driver~\cite{mao2023gpt}. Even with such partial information about the environment, ~\method{} can match or beat the SOTA models in predicting L2 distance from ground truth trajectory.}
\label{tab:nuscenes_long}
\end{table*}

\begin{figure*}[t]
\begin{center}
\centerline{\includegraphics[width=1.\linewidth]{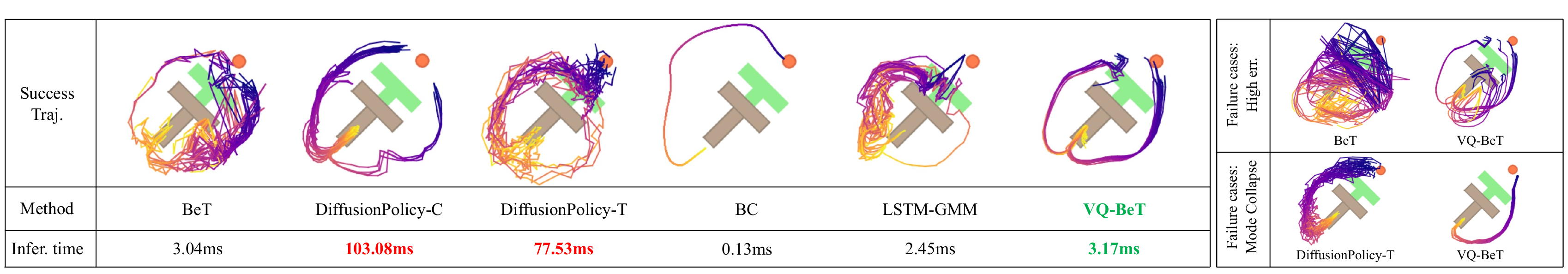}}
\vskip -0.in
\caption{Multi-modal behavior visualization on pushing a T-block to target. On the left, we can see trajectories generated by different algorithms and their inference time per single step, where VQ-BeT generate smooth trajectories to complete the task with both modes with short inference time. On the right, we can see failure cases of VQ-BeT and related baselines due to high error and mode collapse. }
\label{fig:visual-multimodality}
\end{center}
\end{figure*}

\begin{figure}[ht!]
\begin{center}
\centerline{\includegraphics[width=1.\linewidth]{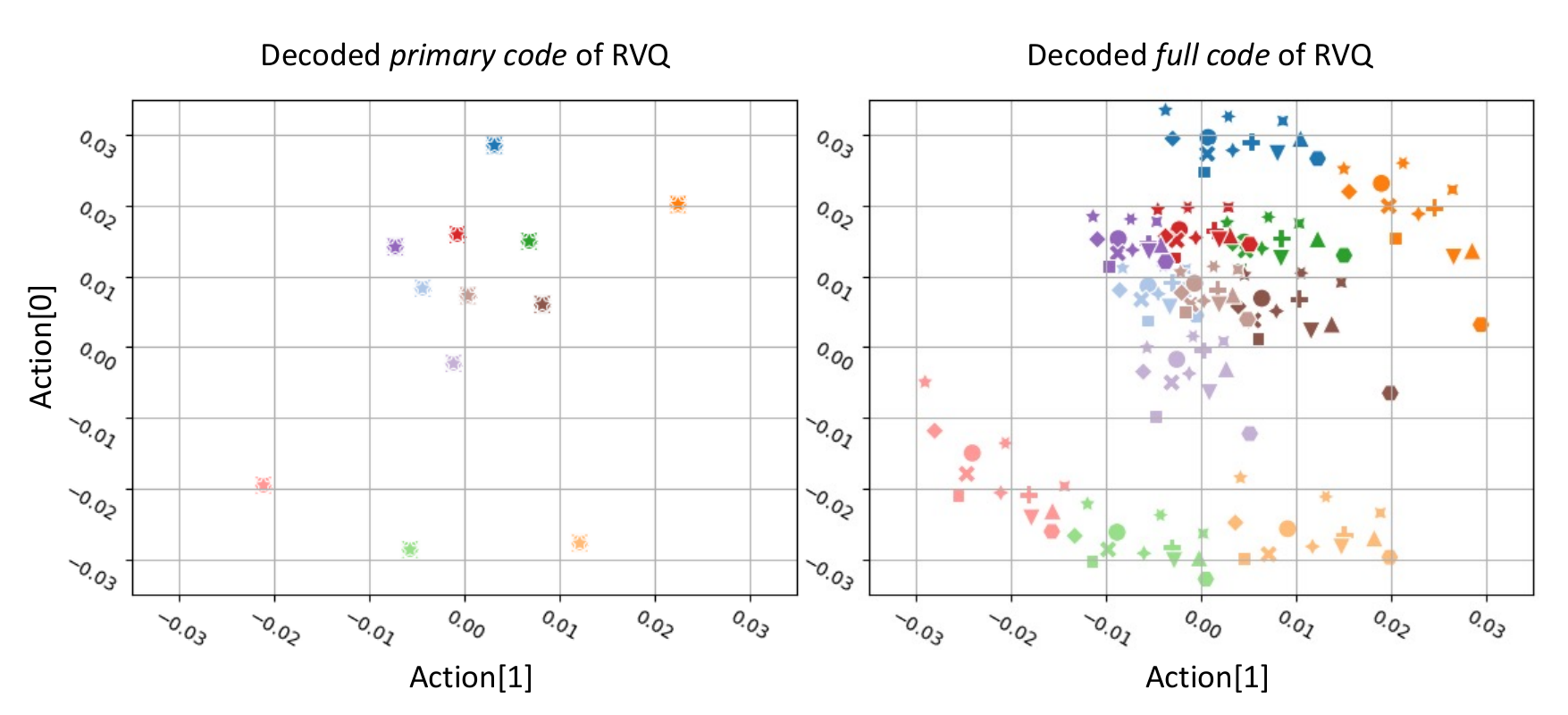}}
\vskip -0.in
\caption{Action centroids of primary codes and full combination of the codes. On the left, we represent centroids of the raw action data obtained by decoding (total of 12) primary codes learned from Blockpush Multimodal dataset. On the right, we show the decoded action of the centroids corresponding to all 144 possible combinations of full the codes. We can see that the primary codes, represented by different colors in each figure, are responsible for clustering in the coarse range, while full-code representation provides further finer-grained clusters with secondary codes.}
\label{fig:decoded_centroids}
\end{center}
\end{figure}

\begin{figure*}[t!]
\begin{center}
\centerline{\includegraphics[width=1.\linewidth]{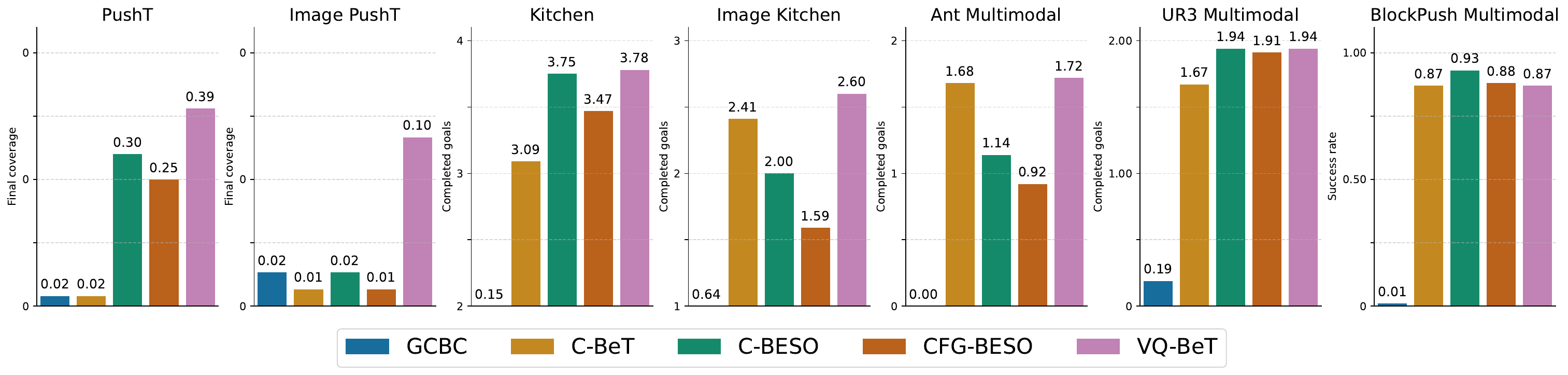}}
\vskip -0.in
\caption{Evaluation of conditional tasks in simulation environments of VQ-BeT and related baselines. VQ-BeT achieves the best performance in most simulation environments and comparable performance with the best baseline on BlockPush.}
\end{center}
\end{figure*}

\begin{figure*}[t!]
\begin{center}
\centerline{\includegraphics[width=1.\linewidth]{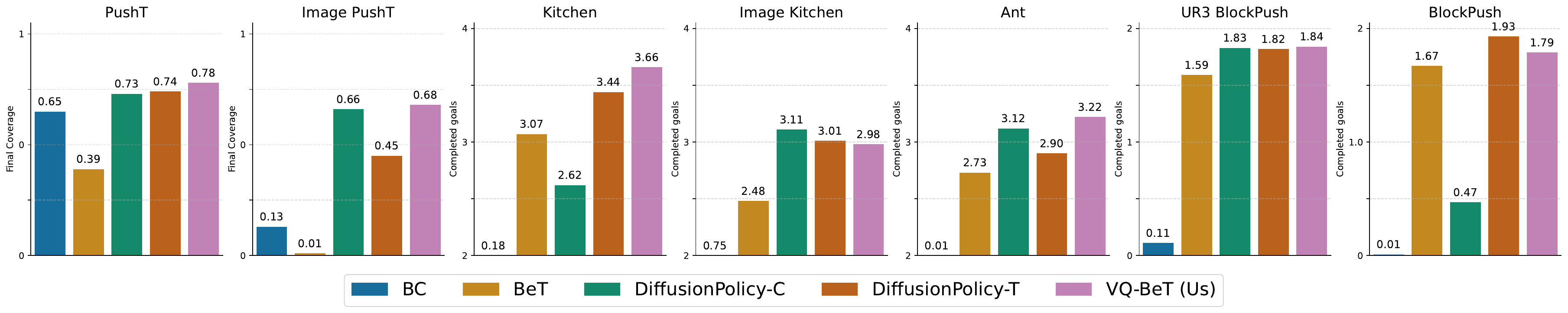}}
\vskip -0.in
\caption{Evaluation of unconditional tasks in simulation environments of VQ-BeT and related baselines. VQ-BeT achieves the best performance in most simulation environments and comparable performance with the best baseline on BlockPush and Image Kitchen.}
\end{center}
\end{figure*}

\begin{table}[ht!]
    \centering
    \resizebox{\textwidth}{!}{%
        \begin{tabular}{@{}lccccccc@{}}
            Control method & Close Toaster & Close Fridge & Can to Toaster & Can to Fridge &
            \begin{tabular}[c]{@{}c@{}}Can to Fridge → \\ Close Fridge\end{tabular} &
            \begin{tabular}[c]{@{}c@{}}Close Fridge \\ and Toaster\end{tabular} & Total \\ 
            \midrule
            Closed loop ($n=1$) & $9/10$ & $8/10$ & $10/10$ & $10/10$ & $4/10$ & $6/10$ & $47/60$ \\
            Receding horizon ($n=3$) & $0/5$ & $0/5$ & $0/5$ & $0/5$ & $0/5$ & $0/5$ & $0/30$ \\
            \bottomrule
        \end{tabular}%
    }
    \caption{Quantitative results of running diffusion policy~\cite{chi2023diffusion} with closed-loop vs. receding horizon control in real-world robot experiments, where $n$ is the number of actions executed at each timestep. We select four single-phase tasks and two two-phase tasks in which diffusion policy does well with closed-loop control, and compare with the same policy with receding horizon control by executing multiple predicted actions at each timestep. We see the diffusion policy with an action sequence executed per timestep goes out of distribution quite easily and fails to complete any tasks on this set of experiments.}
    \label{tab:no-receding-control}
\end{table}

\clearpage
\subsection{VQ-BeT with larger Residual VQ Codebook}

\begin{table*}[ht]
\centering
\scriptsize
\begin{tabular}
{c|c|c|ccc|cccc}
\noalign{\smallskip}\noalign{\smallskip}
\Xhline{2\arrayrulewidth}

\multicolumn{2}{c|}{}& Original codebook & Extended Codebook & Extended Codebook \\
\multicolumn{2}{c|}{}& (Vanilla VQ-BeT) & (Vanilla VQ-BeT) & (VQ-BeT $+$ Deadcode Masking) \\

\hline

\multirow{5}{*}{Ant Multimodal (Unconditional)} & Codebook Size & 10 & 32 & 32\\
                      & $\#$ of Code Combinations  & 100 & 1024 & 1024\\
                      & $(\cdot / 4)$ & \textbf{3.22} & 3.01 & 3.11 \\
                      & p3-Entropy & 4.19 & 4.23 & \textbf{4.33}\\
                      & p4-Entropy & 4.20 & 4.24 & \textbf{4.32}\\
                      \hline

\multirow{3}{*}{Ant Multimodal (Conditional)} & Codebook Size & 10 & 48 & 48\\
                      & $\#$ of Code Combinations  & 100 & 2304 & 2304\\
                      & $(\cdot / 2)$ & 1.72 & 1.75 & \textbf{1.81}\\
                      \hline

\multirow{5}{*}{Kitchen (Unconditional)} & Codebook Size & 16 & 64 & 64\\
                      & $\#$ of Code Combinations  & 256 & 4096 & 4096\\
                      & $(\cdot / 4)$ & 3.66 & \textbf{3.75} & 3.7 \\
                      & p3-Entropy & \textbf{3.42} & 3.01 & 3.10\\
                      & p4-Entropy & \textbf{4.07} & 3.57 & 3.74\\
                      \hline

\multirow{3}{*}{PushT (UnConditional)} & Codebook Size & 16 & 64 & 64\\
                      & $\#$ of Code Combinations  & 256 & 4096 & 4096\\
                      & Final Coverage & 0.78 & 0.77 & \textbf{0.79}\\
                      & Max Coverage & 0.80 & 0.80 & \textbf{0.82}\\
                      \hline

\multirow{3}{*}{Kitchen (Conditional)} & Codebook Size & 16 & 256 & 256\\
                      & $\#$ of Code Combinations  & 256 & 65536 & 65536\\
                      & $(\cdot / 4)$ & \textbf{3.78} & 3.61 & 3.56\\
                      \hline
\end{tabular}
\caption{Evaluation of conditional and unconditional tasks in simulation environments of VQ-BeT with extended size of Residual VQ codebook.}
\label{tab:larger_rvq_dictionary}
\end{table*}

In this section, we present additional results to evaluate the performance of VQ-BeT with larger residual VQ codebooks. While the results of VQ-BeT across the manuscript were obtained using 8 to 16-sized codebooks, resulting in 64 to 256 code combinations (Table \ref{table:hyperparameter_vq_beT}), here, VQ-BET was trained on codebooks with 10 to 250 times more combinations, as detailed in Table \ref{tab:larger_rvq_dictionary}. First, we evaluate VQ-BeT with extended codebook size without any modifications (`Vanilla VQ-BeT'). Next, we test VQ-BeT with an additional technique where the code combinations that do not appear in the dataset are masked with a probability of zero at sampling time to eliminate the possibility of these combinations.

As shown in Table \ref{tab:larger_rvq_dictionary}, we find that increasing the number of combinations ($\times 10 \sim \times 250$) had little impact on performance in most environments. In environments Ant Multimodal (Conditional) and PushT (Unconditional), overall performance slightly increased as the size of the VQ codebook increased. In environments Ant Multimodal (Unconditional) and Kitchen (Unconditional), we see that there is a performance and entropy trade-off as the size of the codebook increases. The only environment where the performance of VQ-BeT decreased with the extended size of the codebook was Kitchen (Conditional). Also, we see that there is no consistent evidence on whether using masking the deadcode (code combinations that do not appear in the dataset) is better: in Ant and PushT environments, masking led to similar or better performance, while in the Kitchen environment, we find similar or slightly worse performance with masking.

Overall, we conclude that VQ-BeT has robust performance to the size of the codebook if it is enough to capture the major modes in the dataset. We conjecture that this robustness is due to VQ-BeT assigning appropriate roles between primary and secondary codes as the codebook size increases. For example, in the Kitchen (Conditional) environment where we have increased the number of possible combinations by 256, the code prediction accuracy rate has decreased by only $\times0.08$ of its original accuracy rate, while the primary code prediction retained $\times0.8$ of its original accuracy rate. Interestingly, Despite this large difference, the performance difference between the two is small, around 4.5\% (3.78 vs 3.61). 
These results suggest that VQ-BeT could rely on the resolution of the primary code in large VQ codebook size, while using less weight on the secondary code to handle the excessive number of code combinations, leading to robust performance to the size of the codebook.

\clearpage

\section{Implementation Details}
\subsection{Model Design Choises}

\begin{table}[h!]
\centering

\resizebox{\textwidth}{!}{%
\begin{tabular}{c|ccccccc}
\hline
Hyperparameter & Kitchen & Ant & BlockPush & UR3 & PushT & NuScenes & Real-world \\
\hline

Obs window size & 10 & 100 & 3 & 10 & 5 & 1 & 6 \\
Goal window size (Conditional Task) & 10 & 10 & 3 & 10& 5 & 1 & - \\
Predicted act sequence length & 1 & 1 & 1 & 10 & 5 & 6 & 1 \\
Autoregressive code pred. & False & False & False & False & False & True & True \\
$\beta$ (Eq.~\ref{eq:code_loss}) & 0.1 & 0.6 & 0.1 & 0.1 & 0.1 & 0.1 & 0.5 \\
\hline
Training Epoch & 1000 & 300 & 1500 & 300 & 2000 & 1000 & 600 \\
Learning rate & 5.5e-5 & 5.5e-5 & 1e-4 & 5.5e-5 & 5.5e-5 & 5.5e-5 & 3e-4 \\
\hline
MinGPT layer num & 6 & 6 & 4 & 6 & 6 & 6 & 6 \\
MinGPT head num & 6 & 6 & 4 & 6 & 6 & 6 & 6 \\
MinGPT embed dims & 120 & 120 & 72 & 120 & 120 & 120 & 120 \\

\hline
VQ-VAE latent dims & 512 & 512 & 256 & 512 & 512 & 512 & 512 \\
VQ-VAE codebook size & 16 & 10 & 8 & 16 & 16 & 10 & 8/10/16 \\
\hline
Encoder (Image env) & ResNet18 & - & - & - & ResNet18 & - & HPR \\
\hline
\end{tabular}
}
\caption{Hyperparameters for VQ-BeT}
\label{table:hyperparameter_vq_beT}
\end{table}

\subsection{VQ-BeT for Driving Dataset}
\label{app:sec:driving-howto}
While all the other environments reported in this paper have a fixed observation dimension at one timestep, NuScenes driving dataset, as processed in the GPT-Driver paper~\cite{mao2023gpt}, could contain the different number of detected objects in each scene. Thus, we make modification to the input types of VQ-BeT to train VQ-BeT with NuScenes driving dataset in response to this change in dimensionality of the obeservation data. The tokens we pass to VQ-BeT are as shown below:

\begin{itemize}
    \item{\textbf{Mission Token} indicates the mission that the agent should follow: go forward / turn left / turn right}
    \item{\textbf{Ego-state Token} contains velocity, angular velocity, acceleration, heading speed, and steering angle.}
    \item{\textbf{Trajectory History Token} contains ego historical trajectories of last 2 seconds, and ego historical velocities of last 2 seconds.}
    \item{\textbf{Object Tokens} contains perception and prediction outputs corresponding to current position, predicted future position, and one-hot encoded class indicator of each object. There are total of 15 classes. \textcolor{gray}{(`pushable-pullable', `car', `pedestrian', `bicycle', `truck', `trafficcone', `motorcycle', `barrier', `bus', `bicycle-rack', `trailer', `construction', `debris', `animal', `emergency')}}
\end{itemize}

\begin{figure*}[bh!]
\begin{center}
\vskip -0.1in
\centerline{\includegraphics[width=.6\linewidth]{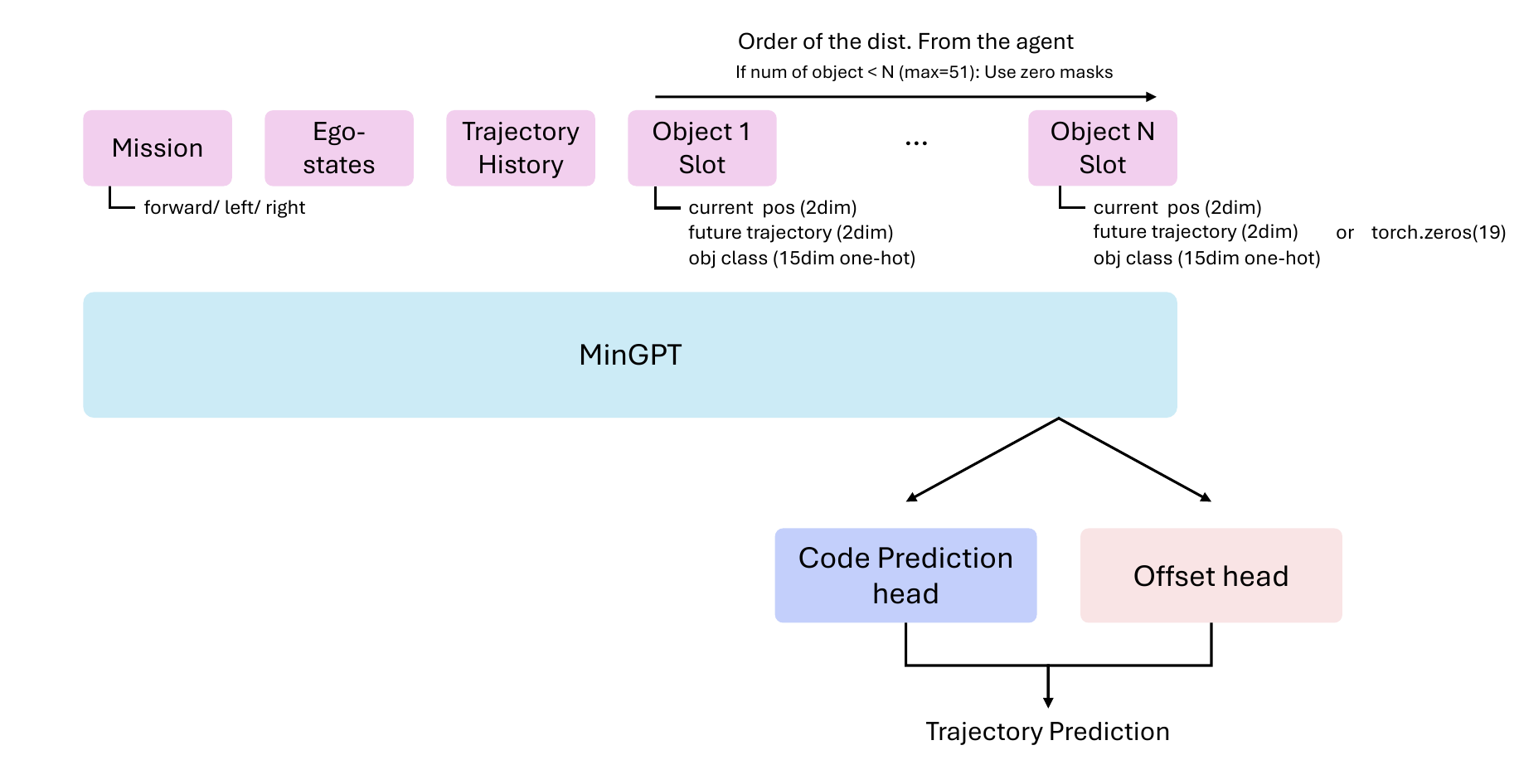}}
\vskip -0.1in
\caption{Overview of VQ-BeT for autonomous driving.}
\end{center}
\vskip -0in
\end{figure*}

\end{document}